\documentclass[journal]{IEEEtran}
\pdfoutput=1

\usepackage{times}
\usepackage{epsfig}
\usepackage{graphicx}
\usepackage{amsmath}
\usepackage{amssymb}
\usepackage{booktabs}
\usepackage{subcaption}
\usepackage{multirow}
\usepackage{diagbox}
\usepackage{xcolor,colortbl}
\usepackage[ruled,linesnumbered]{algorithm2e}
\usepackage{hyperref}
\usepackage{color}
\usepackage{epstopdf}

\ifCLASSINFOpdf
\else
\fi

\begin{document}
%
\title{Learning Enriched Illuminants for Cross and Single Sensor Color Constancy}
%
%
%

\author{Xiaodong~Cun\textsuperscript{*\thanks{\textsuperscript{*}Equal contribution. Work was done during internships at Noak's Ark Lab, Huawei Technologies.}}, Zhendong~Wang\textsuperscript{*},
Chi-Man~Pun, Jianzhuang~Liu, Wengang Zhou, Xu~Jia, Houqiang Li,~\IEEEmembership{Fellow,~IEEE}

\thanks{Xiaodong Cun is with Tencent AI Lab, Shenzhen, 518067, China. E-mail: shadowcun@tencent.com.}%

\thanks{Chi-Man Pun is with Department
of Computer Science and Information, University of Macau, Macau, 999078, China. E-mail: cmpun@umac.mo.}

\thanks{Zhendong Wang, Wengang Zhou and Houqiang Li are with Department of Electronic Engineering and Information Science, University of Science and Technology of China, Hefei, 230026, China. E-mail: zhendongwang@mail.ustc.edu.cn, \{zhwg, lihq\}@ustc.edu.cn.}

\thanks{Jianzhuang Liu is with Noak's Ark Lab, Huawei Technologies, Shenzhen, 518129, China. E-mail: liu.jianzhuang@huawei.com.}

\thanks{Xu Jia is with School of Artificial Intelligence,
Dalian University of Technology, Dalian, 116024, China. E-mail: xjia@dlut.edu.cn.}


}
%
%

\markboth{Journal of \LaTeX\ Class Files,~Vol.~**, No.~*, **~****}%
{Shell \MakeLowercase{\textit{et al.}}: Bare Demo of IEEEtran.cls for IEEE Journals}
%
\def\etc{etc.\@\xspace}
\newcommand{\etal}{\textit{et al.}}

\newcommand*{\eg}{e.g.\@\xspace}
\newcommand*{\ie}{i.e.\@\xspace}
\newcommand{\xxwidth}{0.195}


\maketitle

\begin{abstract}
Color constancy aims to restore the constant colors of a scene under different illuminants. However, due to the existence of camera spectral sensitivity, the network trained on a certain sensor, cannot work well on others. Also, since the training datasets are collected in certain environments, the diversity of illuminants is limited for complex real world prediction. In this paper, we tackle these problems via two aspects.  First, we propose cross-sensor self-supervised training to train the network. In detail, we consider both the general sRGB images and the white-balanced RAW images from current available datasets as the white-balanced agents. Then, we train the network by randomly sampling the artificial illuminants in a sensor-independent manner for scene relighting and supervision. Second, we analyze a previous cascaded framework and present a more compact and accurate model by sharing the backbone parameters with learning attention specifically. Experiments show that our cross-sensor model and single-sensor model outperform other state-of-the-art methods by a large margin on cross and single sensor evaluations, respectively, with only 16\% parameters of the previous best model. 
\end{abstract}


\section{Introduction}




\IEEEPARstart{H}{uman} vision system has a strong ability to perceive relatively constant colors of objects under varying illumination. However, it is difficult for digital cameras to have the same ability. In computer vision, this problem is formulated as computational Color Constancy~(CC), which is also a practical concern for the automatic white balance~(AWB)~\cite{ffcc}.
Besides visual aesthetics, accurate white balance is also necessary for high-level tasks~\cite{Afifi_2019_ICCV}, such as image classification and semantic segmentation.

\begin{figure}[t]
    \centering
    \includegraphics[width=0.9\columnwidth]{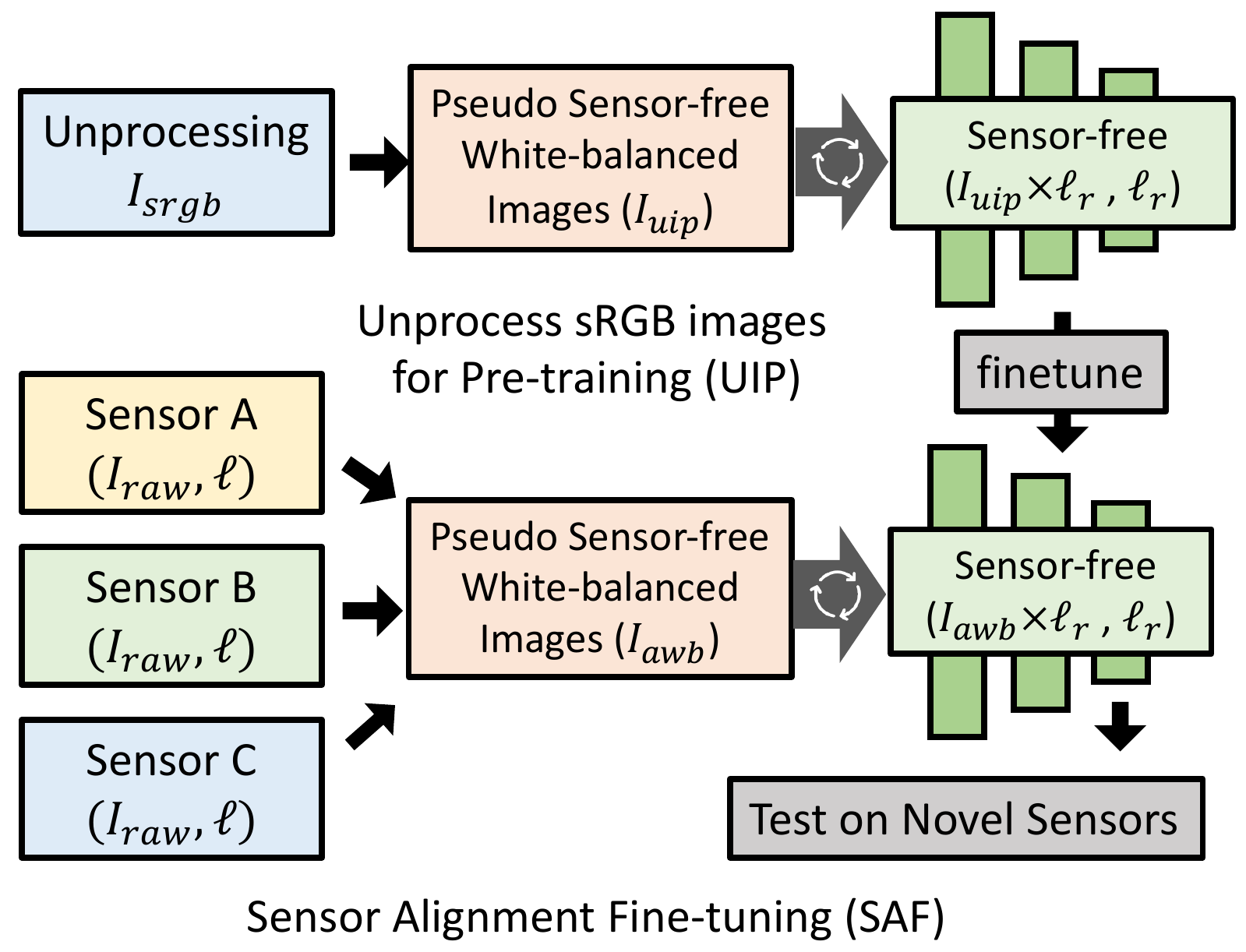}
    \caption{Our method utilizes the white-balanced images for cross sensor CC. Specifically, we consider unprocessed sRGB images~($I_{uip}$) and the corrected RAW images~($I_{awb}$) from real sensors as sensor-free white-balanced images. Then, we utilize these images and randomly sampled illuminants $\ell_{r}$ to train the network according to Eq.~\ref{eq:cc}.}

    \label{fig:xcc}
\end{figure}

In general, a photographed scene image $I_{raw}$ of color channel $c$, $I_{raw}^c$, can be represented as:
\begin{equation}
    I_{raw}^c(x) = \int_{\omega} E(\lambda,x)S(\lambda,x)C_c(\lambda)d\lambda,\quad  c \in \{r, g, b\},
    \label{eq:original}
\end{equation}
where $\omega$ is the visible light spectrum, and $x$ is the spatial location. At wavelength $\lambda$, $C_c(\lambda)$ is the camera sensor sensitivity~(CSS) function, $E(\lambda,x)$ and $S(\lambda,x)$ are the color of light source and the surface spectral reflectance, respectively. Assuming there is only one uniform illuminant in the scene, according to the von Keris model~\cite{von}, the above formula can be simplified as:
\begin{equation}
    I_{raw}^{c} =  I_{awb}^{c} \times \ell^{c} ,\quad c \in \{r, g, b\},
    \label{eq:cc}
\end{equation}
where $I_{awb}^{c}$, and $\ell^{c}$ are the channel $c$ of the corrected white balanced image and the scene illuminant, respectively. In practice, to acquire ground truth illumination for correction or training, a standard object with known chromatic properties~(\eg, color checker~\cite{shi2000re} or SpyderCube~\cite{cube}) often needs to be  placed into the scene. Thus, labelling a large-scale dataset for training a deep neural network is costly. Besides, since there are only 0.2k--0.3k images per sensor in the popular benchmarks~\cite{nus8,shi2000re}, the changes of surface spectral reflectance and color temperature are also insufficient for deep learning. 
Also because of the domain gap between sensors, most previous methods~\cite{hu2017fc4,bianco2015color,ffcc} can only work well for a specific sensor with the models trained on the same sensor.

To overcome aforementioned problems, as shown in Figure~\ref{fig:xcc}, we first propose a novel training technique, called \textit{self-supervised cross-sensor training}, to utilize the knowledge from the images from out-of-domain sensors, and even processed sRGB images. In detail, on the one hand,  we firstly consider the available sRGB datasets as white-balanced because these datasets are carefully corrected by humans or digital cameras. Then, we unprocess the sRGB images and sample the sensor-independent illuminants for training. On the other hand, we collect labelled images from several  sensors and correct the images via Eq.~\ref{eq:cc}. And then, for each white-balanced image, a new illuminant is randomly sampled to relight the image as supervision. Besides the training technique, we design a compact cascaded network based on C4~\cite{yu2020c4}. Considering each iteration in C4 as a residual distribution, we share the parameters in each iteration but learn the illuminant differences with a proposed \textit{iteration-specific attention module}.
Besides, different from C4 which uses FC4~\cite{hu2017fc4} as the backbone, we design a novel \textit{lightweight head} to replace the head in FC4~\cite{hu2017fc4}, reducing 50\% of the parameters while keeping accuracy. 

Our main contributions are summarized as follows:

\begin{itemize}
\item We propose \textit{self-supervised cross-sensor training} to obtain scene knowledge from abundant sRGB images and other available labelled RAW datasets. Once trained using randomly generated labels, our model achieves state-of-the-art performance on cross-sensor evaluation.

\item We design a novel and compact cascaded network structure via an \textit{iteration-specific attention module} and a \textit{lightweight multi-scale head}. 


\item Our single-sensor model achieves state-of-the-art performance on multiple popular datasets using only 16\% of the parameters of the existing best model~\cite{yu2020c4}.

\end{itemize}

\section{Related Work}
Existing methods on color constancy can be approximately categorized into two classes: traditional statistics-based and deep learning-based methods. Below, we give the detailed reviews of each part, while focusing on the discussions about recent works based on deep learning.

\subsection{Traditional Color Constancy}
Early methods estimate the scene illuminant using statistics features~\cite{bright-pixels,gray-index,gray-world, grey-pixel,white-patch,shades-of-gray} of the input image. Based on the assumption that the scene color is achromatic, the mean of each color channel~\cite{gray-world}, the gray edge~\cite{gray-edge} and the white patch~\cite{white-patch} are commonly used features for illumination estimation. Although these methods are simple and fast, the accuracy is unsatisfactory in real world prediction. To further boost the performance in real world scenes, machine learning-based methods show significantly better results than statistics-based methods. A branch of early works extract hand-crafted features~\cite{regression-tree,SVR} for learning, while the most recent works are all based on deep learning, which exhibits great potential in many computer vision tasks~\cite{alexnet,squeezenet,deng2009imagenet} including CC.

\subsection{Deep Color Constancy}

\textbf{Illumination Regression and Classification.} Since CC only needs to predict a global illumination, regression-based methods are easy to be applied. Bianco~\etal~\cite{bianco2015color} propose a patch-based network inspired by image classification~\cite{alexnet}. Then, Qiu~\etal~\cite{qiu2020color} design a feature re-weighted layer and use hierarchical features for regression. Besides a single patch, Xu~\etal~\cite{end2end} involve the triplet loss to constrain the differences among the patches from a single image. However, although illuminant can be simply regarded as a global feature, the low-level features, such as the white patch and gray edge, are also important. Besides regression, considering CC as a classification problem also draws attention from researchers~\cite{oh2017approaching,dsn,multi-hyp}. For example, \cite{multi-hyp} selects multiple candidates and weights them under a Bayesian framework. However, the selection of suitable candidates from the dataset hugely influences the performance and the accuracy is unsatisfactory.

\textbf{White Patch~(Point) Localization.} 
White patch has been shown to be highly connected to CC in statistics-based methods. Some works~\cite{ccc,ffcc,siie} extract a color $uv$-histogram in the log-chromatic space and then localize it following object detection. Differently, some recent fully convolutional neural network-based methods~\cite{hu2017fc4,yu2020c4} can also be considered as white-point localization ones. They extract features by neural networks directly using stronger backbones pre-trained on ImageNet~\cite{deng2009imagenet}. With the backbones, these methods have their novel structures, such as the confidence-weighted pooling layer~\cite{hu2017fc4}, to focus on the most related region automatically. Our proposed network also belongs to this category, but it obtains a more satisfactory accuracy with much fewer parameters.

\textbf{Pre-Training.} Unsupervised especially self-supervised learning draws extensive attention recently, which has been successfully utilized in image classification~\cite{moco,simclr} and natural languge processing~\cite{bert}. In CC, Lou~\etal~\cite{lou2015color} learn to minimize the loss between the network prediction and the labels obtained by a traditional statistic-based method on sRGB images. However, the results of the statistic-based method are inaccurate. Based on the fact that available sRGB images can be considered as white-balanced, Bianco and Cusano~\cite{bianco2019quasi} train their network to detect the achromatic patches by an achromatic loss. However, the heavy network and unsatisfactory performance restrict the usage of their method. Differently, our pre-training aims to regress the illuminant directly, which is simple and effective. Moreover, our pre-trained model can be also used for other state-of-the-art CC networks directly.


\textbf{Cross Sensor CC.}
The existence of CSS is the main reason why a model trained on one particular sensor is hard to be applied to another. Considering CSS, Gao~\etal~\cite{css} rectify the images from a source sensor and learn a model trained on a target sensor by mapped samples. Afifi~\etal~\cite{siie} design a gamut mapping network to map the gamut of inputs into the canonical gamut, which makes it sensor-independent, and estimate the illuminant from the latter gamut. The multi-hypothesis strategy~\cite{multi-hyp} aims at the cross-sensor evaluation by selecting the illuminant candidates on the target sensor data. Also, few-shot meta-learning~\cite{mcdonagh2018formulating} and multi-domain learning~\cite{mdlcc} are applied to CC to squeeze the gap among different domains. Differently, our method leverages the knowledge from larger-scale datasets by randomly sampling illuminants. 

\textbf{Compact Networks.}
Applying current deep learning models on resource-limited devices, such as mobile phones \cite{cheng2017survey,howard2017mobilenets,zhang2018shufflenet}, is a hot topic in the research community. In CC, although many previous networks~\cite{cm,ccc,ffcc,qiu2020color} are designed for lightweight structures and real-time prediction, the applicability of these methods is restricted by either a 
hand-crafted color $uv$-histogram~\cite{ccc,ffcc} or not-good-enough performance~\cite{cm,qiu2020color}. Differently, the proposed model is fully convolutional and obtains the state-of-the-art performance while using relatively fewer parameters.

\section{Method}

Traditionally, CC has widely been explained as a supervised learning problem~\cite{hu2017fc4,yu2020c4}. 
Given a neural network $\psi(I_{raw}; \omega)$ with a raw image $I_{raw}$ as input, the network only predicts a globally three-dimensional normalized vector $\ell_{pred}$ representing the chromatism caused by corresponding illumination. Then, under the supervision of the ground-truth label~$\ell$ from manual acquisition, the parameters of the network $\omega$ are optimized by gradient descent. To simplify the notation in our analysis, we represent the basic training pair~(input,~label) for the white-balanced image $I_{awb}$ as:
\begin{equation}
    I_{awb} \Leftarrow (I_{raw},\ell).
    \label{eq:awb}
\end{equation}
In this work, we find that the diversity of illuminants in datasets is the key to train a cross-sensor CC model and to improve the model performance for single-sensor CC. We propose a method to learn enriched illuminants for cross and single sensor CC. 
Firstly, we propose a self-supervised training technique to train a model for cross-sensor CC~(Sec.~\ref{sec:ss}). Secondly, the trained cross-sensor model can be further fine-tuned on the novel single sensor using the sensor-aware illuminant enhancement~(Sec.~\ref{sec32}).
Thirdly, as for the model design, we utilize the cascaded structure as a relighting-based label augmentation method which is different from C4~\cite{yu2020c4}, and propose a more compact and accurate model~(Sec.~\ref{sec:ccn}). 
Finally, the loss function we used is discussed in Sec.~\ref{sec:loss}.
Below, we give the detail of each part.

\subsection{Self-Supervised Cross-Sensor Training}
\label{sec:ss}


From Eq.~\ref{eq:original}, the white-balanced image is highly related to the CSS~\cite{siie,css,csswacv}, which describes the spectral response caused by a sensor and is typically different between sensors. Also, because of the existence of CSS, collecting a common large-scale labelled dataset for universal~(cross-sensor) CC is unrealistic. Besides, since the datasets are often collected in certain environments, the diversity of the illuminants is also limited~\cite{multi-hyp,oh2017approaching}. 
Thus, leveraging large-scale \textit{sensor-free} knowledge to 
train a deep network is important and necessary in CC. 

Inspired by recent development of self-supervised learning in image colorization~\cite{zhang2016colorful}, in-painting~\cite{deepfill} and machine translation~\cite{bert}, we propose a novel self-supervised cross-sensor training technique including two steps to tackle above issues. 
In detail, we utilize the white-balanced samples generated by unprocessing sRGB images to the RAW domain or correcting RAW images with corresponding labels from multiple sensors. And then we train a cross-sensor model with these images and randomly sampled illuminants. Below, we give the details of each step.

\begin{figure}[t]
    \centering
    \includegraphics[width=0.9\columnwidth]{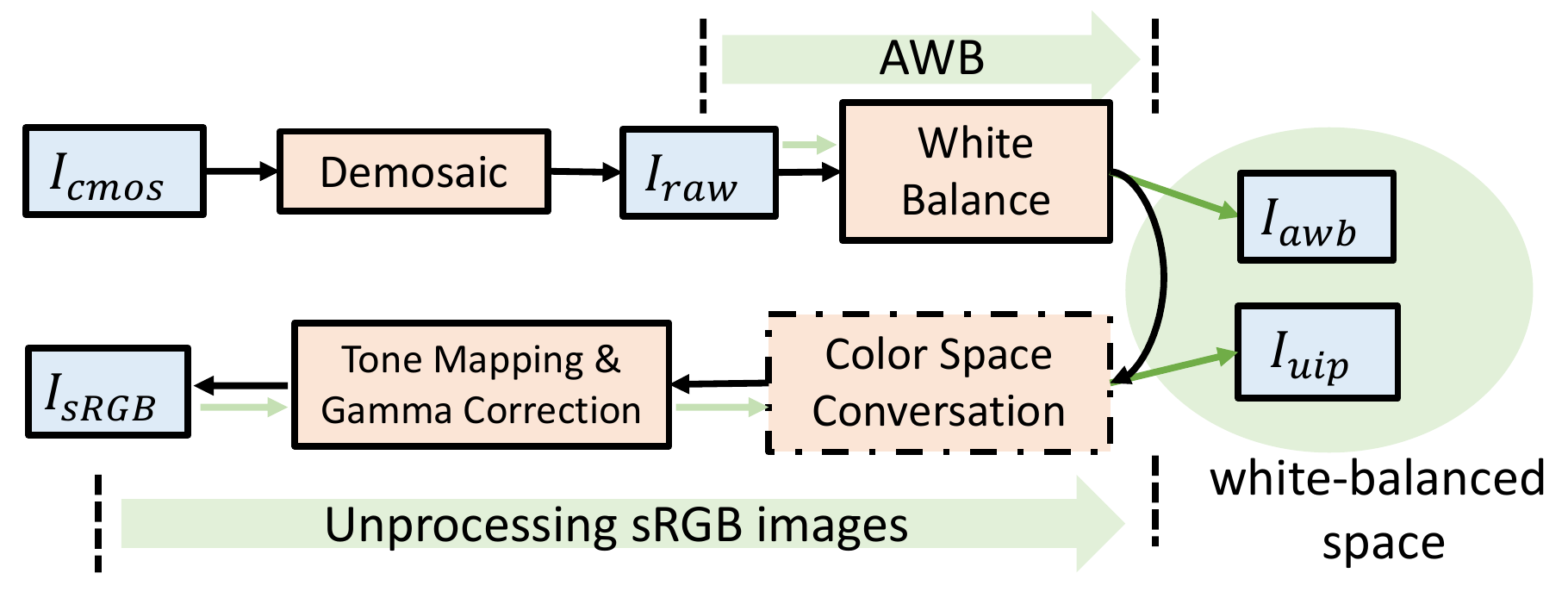}

    \caption{The relation among $I_{raw}$, $I_{awb}$, $I_{uip}$ and $I_{srgb}$ images. After AWB, $I_{raw}$ becomes $I_{awb}$. $I_{uip}$ is obtained by unprocessing an sRGB image~$I_{srgb}$, and is also white-balanced.}

    \label{fig:homo}
\end{figure}

\textbf{Unprocessing sRGB Images for Pre-Training~(UIP).} Since there are millions of well-processed sRGB images available ~\cite{deng2009imagenet,coco}, we treat these sRGB images as white-balanced and unprocess \cite{unprocessing} them to the RAW domain. Differently, since we consider the processed sRGB images are sensor-independent and white-balanced, we only do the inverse tone mapping and de-gamma correction to avoid the color change. Figure~\ref{fig:homo} shows the relation between an sRGB image $I_{srgb}$ and its unprocessed version $I_{uip}$. Then, we regard these images $I_{uip}$ as sensor-free white-balanced samples and randomly sample illuminants within a large range for generating training data. For example, since most of RAW images are in Bayer pattern~($RGGB$), we randomly sample an illuminant $\ell^c_{r} \in [0.2,0.8], c \in \{r,g,b\}$, double the green component and normalize the illuminant as ground truth for each image. Then, a new training combination $I_{uip} \Leftarrow (I_{raw},\ell_{r})$ can be obtained by $I^c_{raw} = I_{uip}^c \times \ell^c_{r}, c \in \{r,g,b\}$, according to Eq.~\ref{eq:cc}. Even if
these generated images may not represent real illuminants in the real world, the proposed method provides an almost unbiased dataset with countless synthesized illuminants for training and involves millions of available sRGB images.

After obtaining a large set of training pairs~(combinations), we train our network with it for the initialization of the network, which will be fine-tuned in the next step SAF.

\begin{figure*}[t]
    \centering
    \includegraphics[width=\textwidth]{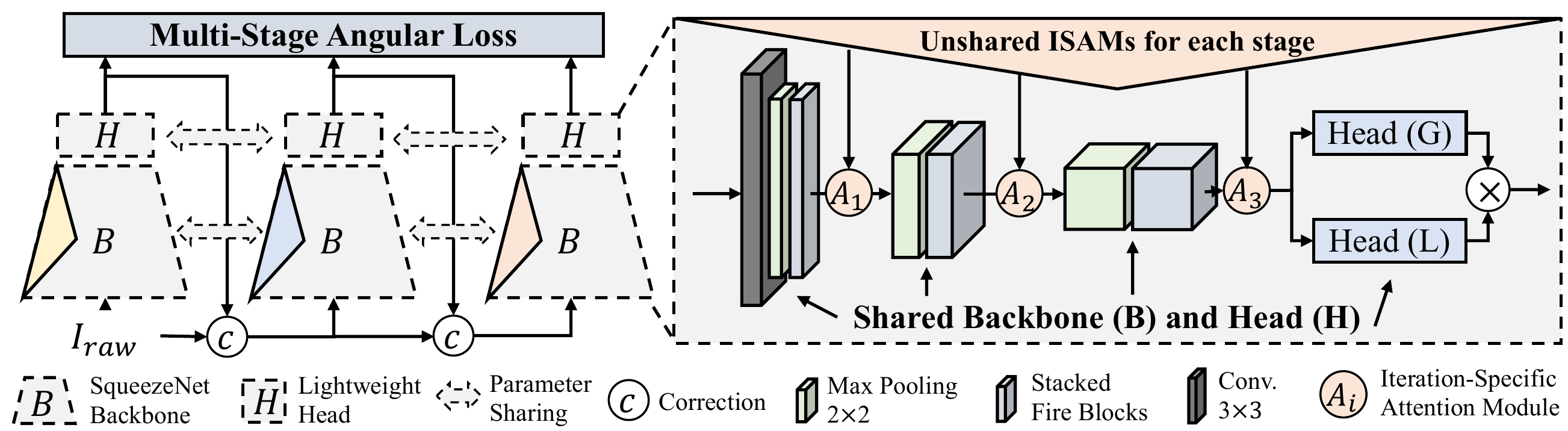}

    \caption{The proposed network structure. The SqueezeNet backbone and the proposed \textit{Lightweight Head} for prediction are shared, and the \textit{Iteration-Specific Attention Module} is learnt specifically in each iteration. The ReLU layer is omitted after each convolutional layer for clarity.}

    \label{fig:nn}
\end{figure*}

\textbf{Sensor Alignment Fine-Tuning~(SAF).} Besides UIP, to match the distribution of real sensors for cross-sensor evaluation, we perform a similar self-supervised method on a dataset including multiple real sensors~\cite{laakom2019intel}. For fine-tuning, we firstly load the network parameters from the previous step UIP for the initialization. Then,
we correct the raw images $I_{raw}$ from each sensor by their corresponding ground-truth illumination labels $\ell$ using Eq.~\ref{eq:cc} to generate the sensor-free white-balanced images $I_{awb}$. Finally, we fine-tune the network with the samples and randomly sampled illuminants. Experiments show that our fine-tuned model can work well for \textit{novel sensor evaluation without fine-tuning on specific sensor data}.

\subsection{Sensor-Aware Illuminant Enhancement for Single-Sensor Training}
\label{sec32}

For single-sensor CC training, we also improve the performance by synthesizing illuminants that follow the distribution of the sensor's CSS similar to \cite{fourure2016mixed,lou2015color,hu2017fc4}. In detail, we first correct the RAW images $I_{raw}$ by their corresponding ground-truth illumination labels $\ell$ according to Eq.~\ref{eq:cc}. Then, new training samples are generated by randomly selected illuminants $\ell^\dag$ from the available data of the \textit{corresponding sensor}, which can be formulated as: $ \frac{I_{raw}}{\ell} \times \ell^\dag =  I_{awb} \times \ell^\dag,$ where we can train the network using the new combinations $I_{awb} \Leftarrow ( \frac{I_{raw}}{\ell} \times \ell^\dag, \ell^\dag)$. We name this scene relighting method as \textit{sensor-aware label reshuffle}~(Reshuffle).

Although Reshuffle can be used solely during training to improve performance for a specific sensor, we further combine random relighting\footnote{Similar to~\cite{hu2017fc4}, we rescale the images and corresponding labels by random RGB values in [0.6, 1.4], and normalize the results.} and the original training pairs to train the network alternatively on the fly for single-sensor CC, because it has been shown that training the network with some noise can also increase performance and stabilize the training process~\cite{sukhbaatar2014training,zhou2019towards}. We call 
our hybrid solution \textit{sensor-aware illuminant enhancement}~(SIE) and give the detailed analysis in the experiments.

\subsection{Compact Cascaded Network}
\label{sec:ccn}
As for the network, we start with the previous work FC4~\cite{hu2017fc4} and C4~\cite{yu2020c4} due to their roles as the backbone and baseline for our work, respectively. In detail, FC4 uses the first 12 layers of the pre-trained SqueezeNet~\cite{squeezenet} as the feature extractor and two convolutional layers with confidence-weighted pooling to estimate the illuminant. Then, C4 cascades FC4 multiple times by correcting the input via Eq.~\ref{eq:cc}. Denoting a single stage C4 with parameters $\omega$ as $\psi(;\omega)$, we formulate the $i$-th iteration as: $\ell'_{i} = \psi_{i}(I_{raw}/\prod_{j = 0}^{i-1}\ell'_{j}; \omega_{i})$, where $\ell'_{i}$ is the output of the $i$-th stage $\psi_{i}$ and $\ell'_{0} = (1,1,1)$. The problem with this structure is that it is too heavy since it requires to optimize and store $i$ times more parameters than FC4, which makes it easy to overfit and far more redundant.




\begin{figure}[t]
    \centering
    \includegraphics[width=\columnwidth]{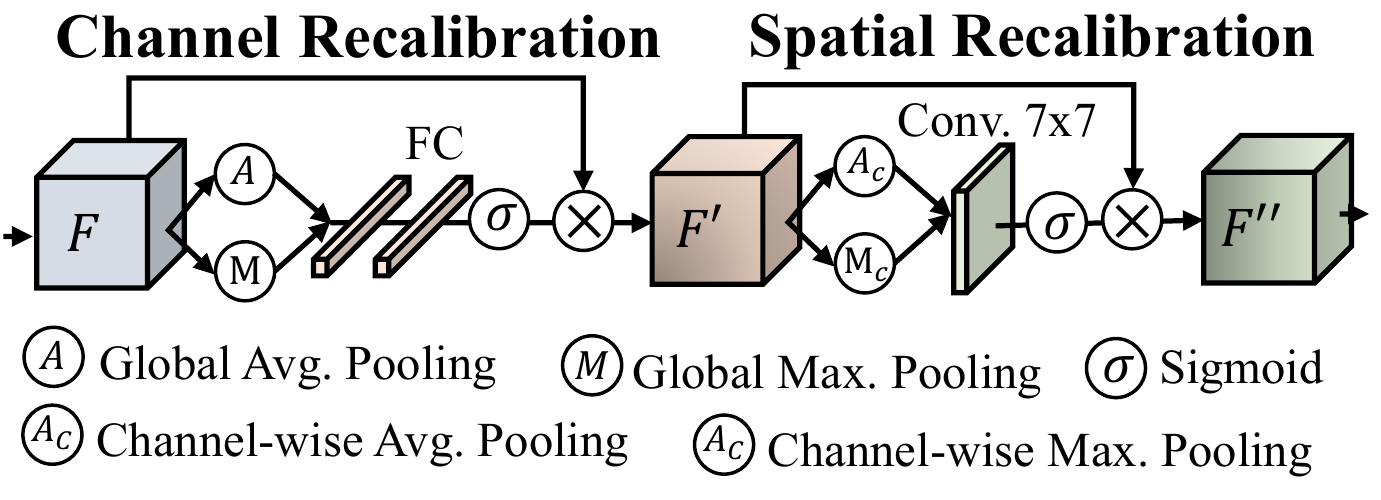}

    \caption{The detailed structure of the proposed Iteration-Specific Attention Module.}

    \label{fig:isam}
\end{figure}

Since all the stages in C4 always learn the same semantic features for the illuminant, and their only difference comes from the percentage among the color channels of the illuminant, we share the backbone in each stage of C4 and design an \textit{iteration-specific attention module} to re-calibrate the difference individually as shown in Figure~\ref{fig:nn}. The proposed framework has several advantages. 
First, sharing the backbone parameters helps us to build a compact model while obtaining better performance than C4. 
Second, since our network contains much fewer parameters, it is easier to be optimized than the original cascaded framework.
Finally, 
the training combinations in different stages~(sub-networks) of C4 are different. For example, the $i$-th sub-network uses this combination $(I_{raw}/\prod_{j = 0}^{i-1}\ell'_{j}, \ell/\prod_{j = 0}^{ i-1}\ell'_{j})$ for training. However, in our shared sub-network, the shared backbone views more information than each of the sub-networks in C4 and its training combination is equivalent to the sum of the $M$-stages' combinations in C4:~$\sum_{i=1}^{M}(I_{raw}/\prod_{j = 0}^{i-1}\ell'_{j}, \ell/\prod_{j = 0}^{i-1}\ell'_{j})$. Obviously, this is a stronger data augmentation of relighting.

We also design a lightweight head for FC4 as our backbone~(L-FC4) which further reduces nearly 50\% of the parameters but gains better performance. Below, we give the details of the proposed \textit{iteration-specific attention module} and \textit{lightweight head}.

\textbf{Iteration-Specific Attention Module~(ISAM).}
 As discussed previously, ISAM aims to learn the color difference between stages using a lightweight structure. As shown in Figure~\ref{fig:isam}, the proposed attention block contains channel re-calibration and spatial re-calibration. In the channel re-calibration, inspired by the previous statistics-based gray-world~\cite{gray-world} and white-patch~\cite{white-patch}, we extract the global per-channel maximum and average values via global max pooling and global average pooling, respectively. Then, these features are learnt via two fully-connected layers. Finally, a sigmoid layer is used to re-calibrate the original features channel-wisely. The spatial re-calibration is similar to the channel re-calibration. Differently, we use channel-wise pooling for the features, and then a $7\times7$ convolutional layer and a sigmoid layer are used for the spatial feature learning and re-calibration. 
 As shown in Figure~\ref{fig:nn}, we insert multiple ISAMs with \textit{different} parameters before each max-pooling layer of the backbone except for the first one. 
The proposed ISAM is inspired by the widely-used attention modules~\cite{seblock,cbam} for image restoration~\cite{zamir2020cycleisp,liu2019dual} and high-level tasks~\cite{seblock,cbam}. However, to the best of our knowledge, for the first time, it is used as the \textit{iteration controller} in the cascaded framework.

\begin{figure}[t]
    \centering
    \includegraphics[width=\columnwidth]{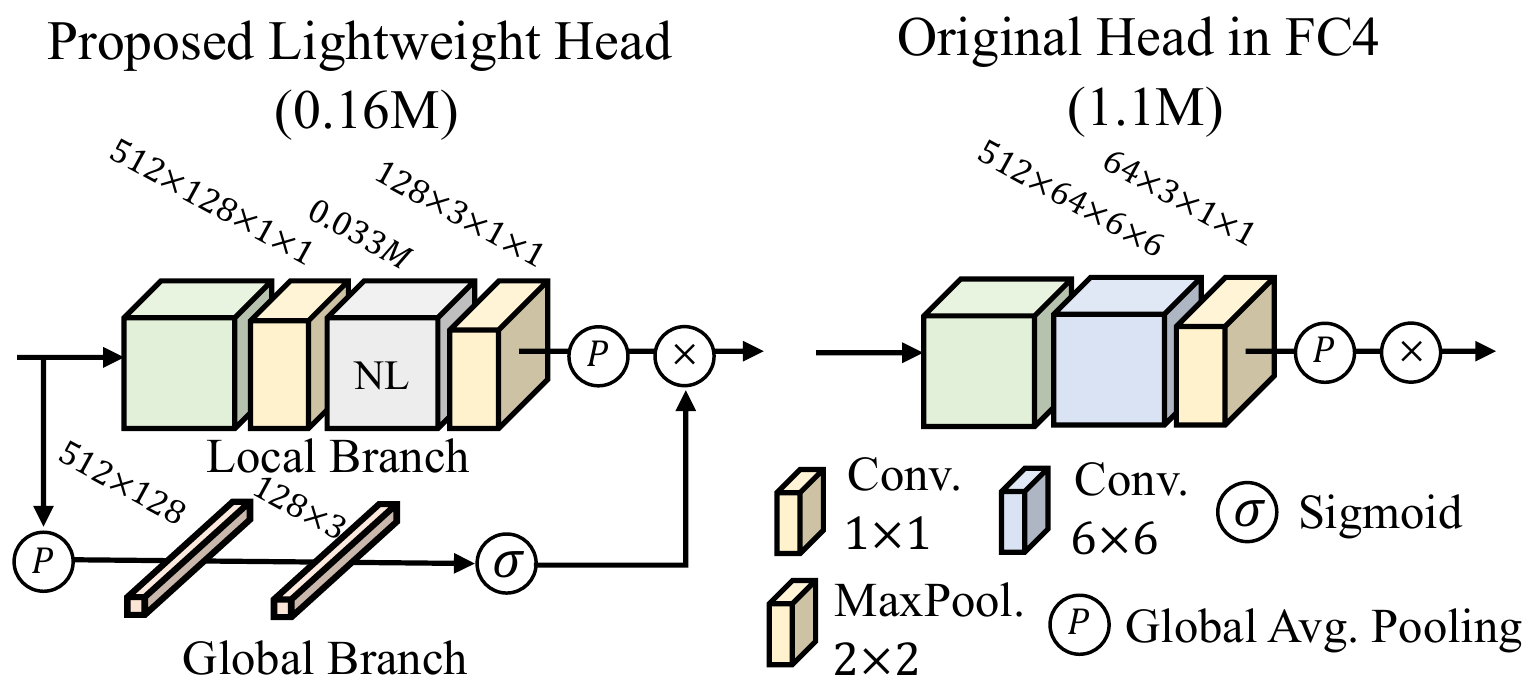}
    \caption{Comparison of parameters between the proposed lightweight head and the original head in FC4. Here, NL  represent Non-Local Block~\cite{non-local}.}
    \label{fig:head}
\end{figure}

\textbf{Lightweight Head.}
As shown in Figure~\ref{fig:head}, we also design a novel lightweight head to get a more compact backbone compared with the original FC4~\cite{hu2017fc4}. Specifically, besides the feature extraction from the first 12 layers of SqueezeNet, FC4 uses two convolutional layers, Conv6~(with $ 512\times64\times6\times6 $ parameters) and Conv7~(with $ 64\times3\times1\times1 $ parameters), to get the final global illumination. The Conv6 is designed to capture larger reception fields on high-level features, which is cumbersome and costs nearly 60\% of the parameters of the overall network. 
Thus, to reduce the parameters and also capture the long-range dependence, we design a lightweight head using a two-branch framework as shown in Figure~\ref{fig:nn}. In detail, in the local~(upper) branch, we reduce the original features spatially by $2\times2$ max pooling and then squeeze the original $512$ channels to $128$ by a $1\times1$ convolutional layer in order to work in a more compact space. Then, we estimate the long-range and global information via a non-local block~\cite{non-local}. After that, a 3-component prediction in the upper branch is generated by a $1\times1$ convolutional layer and global average pooling. We also design a global~(lower) branch by average pooling the original features globally. Then, two linear layers and a sigmoid layer are used to get another 3-component prediction in the lower branch. Finally, the predictions of these two branches are multiplied to generate the final illuminant with $L_2$ normalization for the current stage. The upper branch can be thought as white-patch localization, while the lower branch extracts the global information. This structure shows better performance and captures long-range dependencies as well, which has a negligible amount of parameters added to the backbone. 



\subsection{Loss Function}
\label{sec:loss}
Similar to C4~\cite{yu2020c4}, we use the multi-stage normalized angular loss as the objective. Specifically, for a network with $M$~($M=3$ in ~\cite{yu2020c4} and this paper) iterations~(stages), the  loss $\mathcal{L}$ is defined as:
\begin{equation}
    \mathcal{L} = \sum_{i=1}^{M} \frac{180}{\pi}\arccos({\frac{\ell \cdot \ell'_{i}}{\parallel \ell \parallel \parallel \ell'_{i} \parallel}}),
\end{equation}
where  $\ell'_{i} = \ell'_{i-1} \times \psi_{i}(I_{raw}/\ell'_{i-1})$ is the prediction of the $i$-th stage and $\ell'_{0} = (1,1,1)$ . 

\section{Experiments}
\label{sec:exp}
\begin{table*}[t]
    \small 
    \centering
    \caption{Comparison of angular errors with the state-of-the-art methods on the Gehler-Shi and NUS8 datasets. The \colorbox{lightgray!30}{shadowed} methods are for single-sensor evaluation while the others are for cross-sensor evaluation.}

    \begin{tabular}{l|c|ccccc|ccccc|c}
    \toprule
     \multirow{3}{*}{
     \diagbox{Method}{Angular Error}{Dataset}} & \multirow{3}{*}{Year} & \multicolumn{5}{c|}{Gehler-Shi} &\multicolumn{5}{c|}{NUS8} & \multirow{3}{*}{\#~Param.}\\
    \cline{3-12} 
    & &  Mean & Med. & Tri. & Best & Worst &  Mean & Med. & Tri. & Best & Worst \\
    & &   &  &  & 25\% & 25\% &   &  &  & 25\% & 25\%\\
    \hline
    Grey-World~\cite{gray-world} & 1980 & 6.36 & 6.28 & 6.28 & 2.33 & 10.58 & 4.59 & 3.46 & 3.81 & 1.16 & 9.85 & -\\
    White-Patch~\cite{white-patch} & 1986 & 7.55 & 5.68 & 6.35 & 1.45 & 16.12 & 9.91 & 7.44 & 8.78 & 1.44 & 21.27 & -\\
    SVR~\cite{SVR}& 2004 & 8.08 & 6.73 & 7.19 & 3.35 & 14.89 & - & - & - & - & - & - \\
    Shades-of-Gray~\cite{shades-of-gray}& 2004 & 4.93 & 4.01 & 4.23 & 1.14 & 10.20 & 3.40 & 2.57 & 2.73 & 0.77 & 7.41 & -\\
    1st-order Gray-Edge~\cite{gray-edge}& 2007 & 5.33 & 4.52 & 4.73 & 1.86 & 10.03 & 3.20 & 2.22 & 2.43 & 0.72 & 7.69 & -\\
    2nd-order Gray-Edge~\cite{gray-edge}& 2007 & 5.13 & 4.44 & 4.62 & 2.11 & 9.26 & 3.20 & 2.26 & 2.44 & 0.75 & 7.27 & - \\
    Bayesian~\cite{gehler2008bayesian} & 2008 & 4.82 & 3.46 & 3.88 & 1.26 & 10.49 & 3.67 & 2.73 & 2.91 & 0.82 & 8.21 & - \\
    Quasi-unsupervised~\cite{bianco2019quasi}& 2019 &  3.00 & 2.25 & - & - & - & 3.46 & 2.23 & - & - & - & 54.0M \\
    SIIE~\cite{siie} & 2019 & 2.77 &  1.93  & - & \textbf{0.55} & 6.53 & 2.05 & 1.50 & - & 0.52 & 4.48 & 1.8M \\
    Ours~(cross-sensor) &  2021 & \textbf{2.46} & \textbf{1.89} & \textbf{2.00} & 0.62 & \textbf{5.34} & \textbf{1.98} & \textbf{1.42} & \textbf{1.55} & \textbf{0.48} & \textbf{4.40} & 0.8M \\ \hline
    \rowcolor{lightgray!30} Corrected-Moment~\cite{Corrected-Moment}& 2013 & 2.86 & 2.04 & 2.22 & 0.70 & 6.34 & 2.95 & 2.05 & 2.16 & 0.59 & 6.89 & - \\
    \rowcolor{lightgray!30} Regression Tree~\cite{regression-tree}& 2015 & 2.42 & 1.65 & 1.75 & 0.38 & 5.87 & 2.36 & 1.59 & 1.74 & 0.49 & 5.54 & 31.5M \\
    \rowcolor{lightgray!30} CCC(dist+ext)~\cite{ccc}& 2015 & 1.95 & 1.22 & 1.38 & 0.35 & 4.76 & 2.38 & 1.48 & 1.69 & 0.45 & 5.85 & 0.7K \\
    \rowcolor{lightgray!30} DS-Net(HypNet+SeNet)~\cite{dsn} & 2016 & 1.90 & 1.12 & 1.33 & 0.31 & 4.84 & 2.24 & 1.46 & 1.68 & 0.48 & 6.08 & 5.3M \\
    \rowcolor{lightgray!30} FC4(SqueezeNet-FC4)~\cite{hu2017fc4} & 2017 & 1.65 & 1.18 & 1.27 & 0.38 & 3.78 & 2.23 & 1.57 & 1.72 & 0.47 & 5.15 & 1.7M \\
    \rowcolor{lightgray!30} FFCC(Model M)~\cite{ffcc}& 2017 & 1.78 & 0.96 & 1.14 & 0.29 & 4.62  & 1.99 & 1.31 & 1.43 & \textbf{0.35} & 4.75 & 8.2K \\
    \rowcolor{lightgray!30} Convolutinal-Mean~\cite{cm}& 2019 & 2.48 & 1.61 & 1.80 & 0.47 & 5.97 & 2.25 & 1.59 & 1.74 & 0.50 & 5.13 & 1.1K \\

    \rowcolor{lightgray!30} Multi-Hyp.~\cite{multi-hyp} & 2020 & 2.35 & 1.43 & 1.63 & 0.40 & 5.80 & 2.39 & 1.61 & 1.74 & 0.50 & 5.67 & 22.8K \\
    \rowcolor{lightgray!30} Deep Metric Learning~\cite{end2end}& 2020 & 1.58 & 0.92 & - & 0.28 & 3.70 & 1.85 & 1.24 & - & 0.36 & 4.58 & $>$4.3M \\
    \rowcolor{lightgray!30} C4~\cite{yu2020c4} & 2020 & 1.35 & 0.88 & 0.99 & 0.28 & 3.21  & 1.96 & 1.42 & 1.53 & 0.48 & 4.40 & 5.1M \\ 
   
    \rowcolor{lightgray!30} Ours~(single-sensor) & 2021  &  \textbf{1.27} & \textbf{0.80} & \textbf{0.90} & \textbf{0.24} & \textbf{3.13} & \textbf{1.72} & \textbf{1.23} & \textbf{1.32} & 0.39 & \textbf{3.87} & 0.8M \\
    
    \bottomrule
    \end{tabular}

    \label{tab:compare}
\end{table*}
We conduct extensive experiments on many popular datasets. We start by introducing the basics of the datasets we used and the corresponding experimental settings  in Sec.~\ref{sec:setting}. Then, we demonstrate the comparisons with state-of-the-art methods quantitatively and qualitatively in Sec.~\ref{sec:exp_compare}. Next, the evaluation of the proposed self-supervised cross-sensor training and the single-sensor sensor-aware illuminant enhancement are performed in Sec.~\ref{sec:exp_cross} and Sec.~\ref{sec:exp_single}, respectively. Finally, we provide detailed ablation study of our proposed network structure in Sec.~\ref{sec:exp_net}.

\subsection{Settings}
\label{sec:setting}

\begin{figure}[t]
    \centering
    \includegraphics[width=0.8\columnwidth]{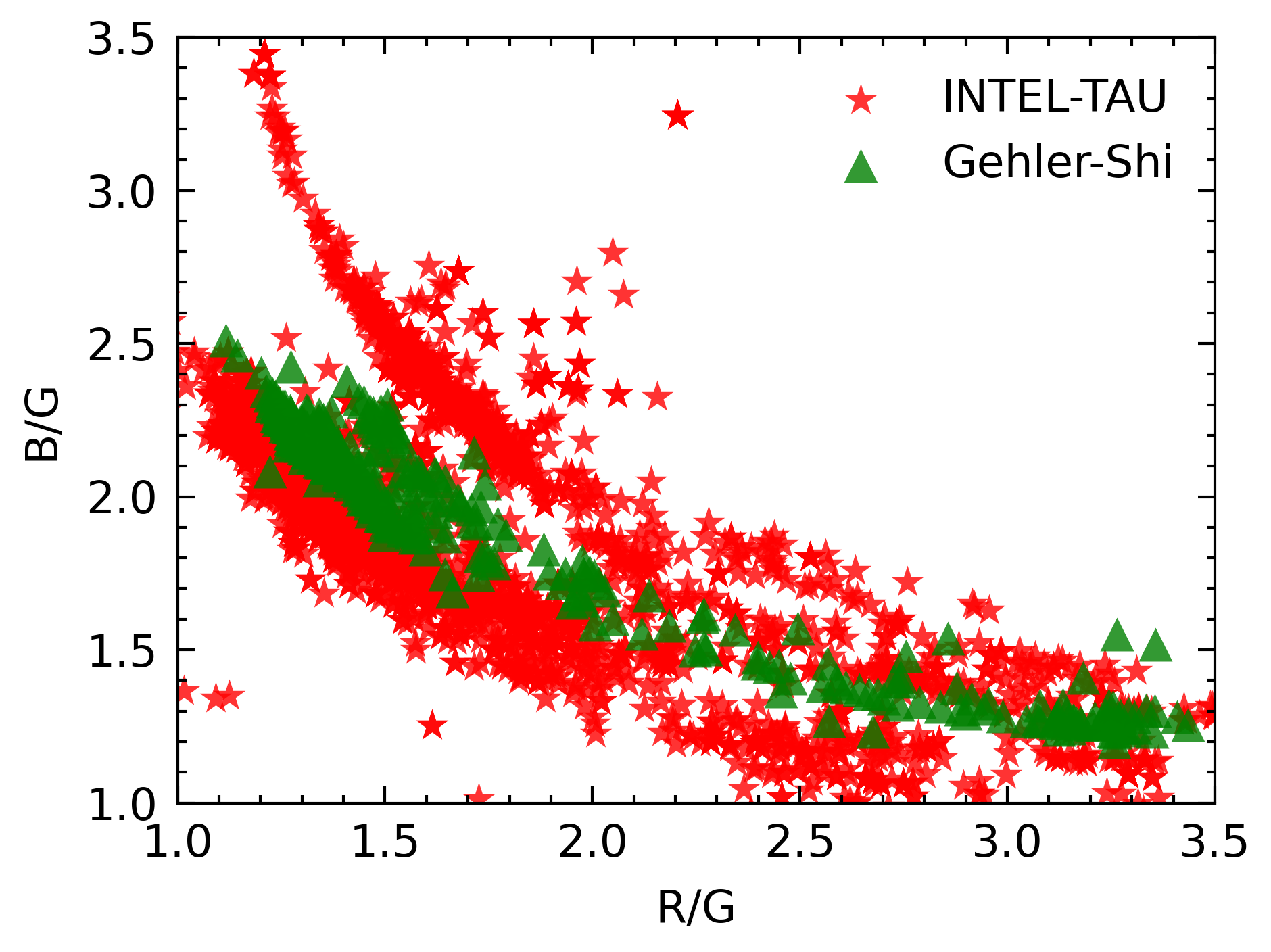}
    \caption{The illuminant distributions in $[\frac{G}{B},\frac{G}{R}]$ of \textit{Canon 5D}~(Gehler-Shi) and \textit{Canon 5DSR}~(INTEL-TAU). Obviously, the two distributions are different.}
    \label{fig:evidence}
\end{figure}

\textbf{Datasets.}
Our method is evaluated on several popular datasets, including the reprocessed Gehler-Shi dataset \cite{gehler2008bayesian,shi2000re}, the NUS 8-Camera dataset~\cite{nus8}~(NUS8), the Cube/Cube+ dataset~\cite{cube} and the Cube+ ISPA 2019 challenge dataset~\cite{cubechallenge}~(Cube-Challenge). 
Specifically, the Gehler-Shi dataset contains 568 images including indoor and outdoor scenes captured by \textit{Canon 1D}~(86 images) and \textit{Canon 5D}~(482 images) cameras. The NUS8 dataset consists of 1736 images from 8 cameras, about 210 images per camera, and each scene may be photographed by multiple cameras. The Cube+ dataset extends the original 1365 images in the Cube dataset with additional 342 images captured in nighttime outdoor and indoor scenes. And Cube-Challenge includes only 363 images for testing a network trained on Cube+. All images in Cube/Cube+/Cube-Challenge are captured by a \textit{Canon 550D} camera. 
For all images in above datasets, a Macbeth Color Checker or a SpyderCube calibration object is put in the scene to obtain the ground-truth illuminant. To avoid the influence of the calibration objects, they are masked out during both training and testing. Black-level subtraction and saturate-pixel clipping are applied after the masking out~\cite{hu2017fc4,multi-hyp}. As for UIP, we use the split of train2017 from COCO~\cite{coco} to train our network. COCO contains 330K common images, which are considered as white-balanced in our experiments. As for SAF, we use the largest available CC dataset INTEL-TAU~\cite{laakom2019intel}. INTEL-TAU contains over 7000 images which are captured by three sensors, \textit{Canon 5DSR}, \textit{Nikon D810}, and \textit{Sony IMX135}. Notice that, all the camera sensors are not overlapping for our cross-sensor evaluation. Specifically, the Canon 5D~(Gehler-Shi) is different from the Canon 5DSR~(INTEL-TAU) and we give the evidence of their different illuminant distributions in Figure~\ref{fig:evidence}.

\textbf{Data Augmentation and Pre-processing.} When training, we apply some basic data augmentation methods similar to previous works~\cite{hu2017fc4,yu2020c4,mdlcc}. 
In detail, for each training sample, we randomly crop a square patch from the original image with a side length of 0.1--1 times of the shorter dimension of the image. Then, this patch is rotated by a random angle in $[-30^{\circ}, 30^{\circ}]$, and flipped horizontally with the probability of 0.5.
Finally, we get the largest rectangle within the rotated patch and resize it into $512\times512$ resolution for training. 
In testing, we only resize the original images to 50\% of their original resolution for speed-accuracy trade-off like ~\cite{hu2017fc4,yu2020c4}. 
All the samples are pre-processed using a gamma correction factor~($\gamma=\frac{1}{2.2}$) to match the distribution of the ImageNet data that are used to pre-train the backbone.

\textbf{Implementation Details.} We implement our network using PyTorch~\cite{paszke2019pytorch}. It is trained in an end-to-end fashion on an Nvidia Tesla V100 GPU. We train it for 4k epochs~(1k epochs for Cube-Challenge) with a batch size of 16 by the Adam optimizer~\cite{kingma2014adam} with a learning rate $3\times10^{-4}$~(divided by 2 after 2k epochs). We use the same number of stages~($M=3$) as C4. As for inference time, our model runs in real-time~(36 fps on $512\times512$ images) and shows a similar speed to the baseline C4 under the same platform. 


\textbf{Evaluation.} For fair comparison, we perform three-fold cross validations on both the Gehler-Shi and NUS datasets using the same splits as \cite{hu2017fc4,multi-hyp,yu2020c4}. When evaluating on Cube-Challenge, we use Cube+ as the training dataset and test on Cube-Challenge following the challenge guideline~\cite{cubechallenge}. Similar to previous works~\cite{yu2020c4,hu2017fc4,mdlcc}, we use the angular error $\epsilon = \frac{180}{\pi}\arccos({\frac{\ell \cdot \ell'}{\parallel \ell\parallel \parallel \ell' \parallel}})$ to measure the difference between the ground-truth illumination $\ell$ and the estimated illumination $\ell'$. We report five common CC statistics about angular errors, \ie, Mean, median~(Med.), tri-mean~(Tri.) of the errors, mean of the smallest 25\% of the errors~(Best 25\%) and mean of the largest 25\% of the errors~(Worst 25\%). The units of these metrics are degrees.

\begin{figure*}[t]
\centering
\begin{subfigure}{\xxwidth\textwidth}
    \centering
    \includegraphics[width=0.88\columnwidth]{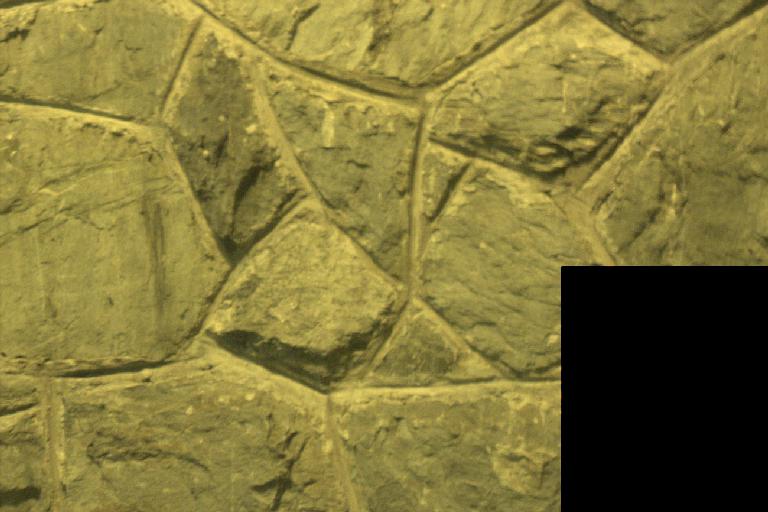}
    \vspace{-0.4em}
    \caption*{Input}
\end{subfigure}
\begin{subfigure}{\xxwidth\textwidth}
    \centering
    \includegraphics[width=\columnwidth]{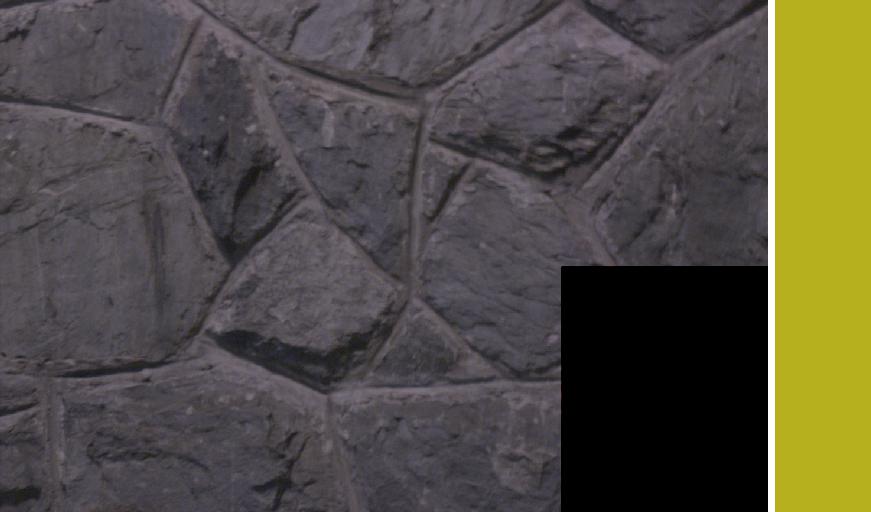}
    \vspace{-1.5em}
    \caption*{FC4~($4.88^{\circ}$)}
\end{subfigure}
\begin{subfigure}{\xxwidth\textwidth}
    \centering
    \includegraphics[width=\columnwidth]{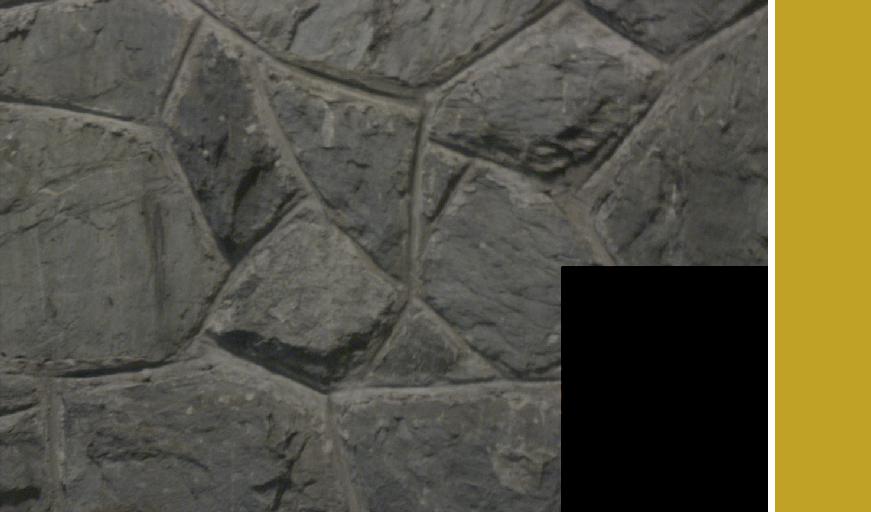}
    \vspace{-1.5em}
    \caption*{C4~($1.05^{\circ}$)}
\end{subfigure}
\begin{subfigure}{\xxwidth\textwidth}
    \centering
    \includegraphics[width=\columnwidth]{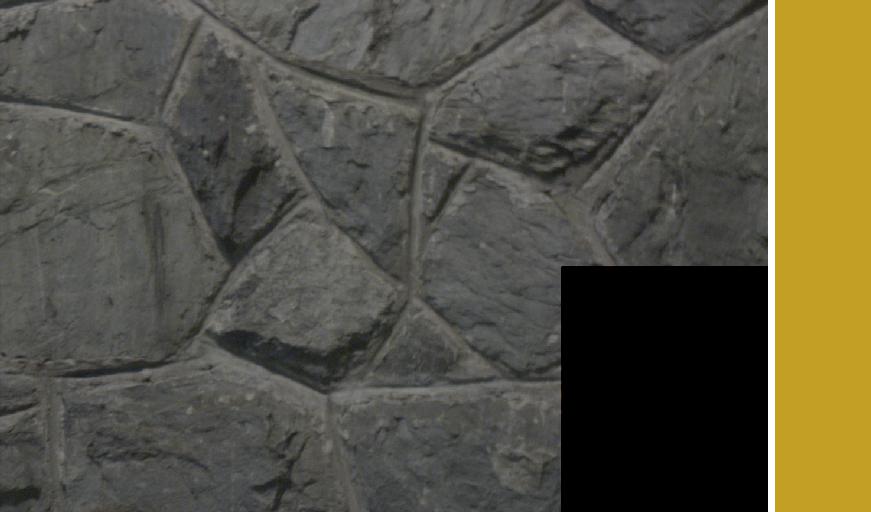}
    \vspace{-1.5em}
    \caption*{Ours~($\mathbf{0.10^{\circ}}$)}
\end{subfigure}
\begin{subfigure}{\xxwidth\textwidth}
    \centering
    \includegraphics[width=\columnwidth]{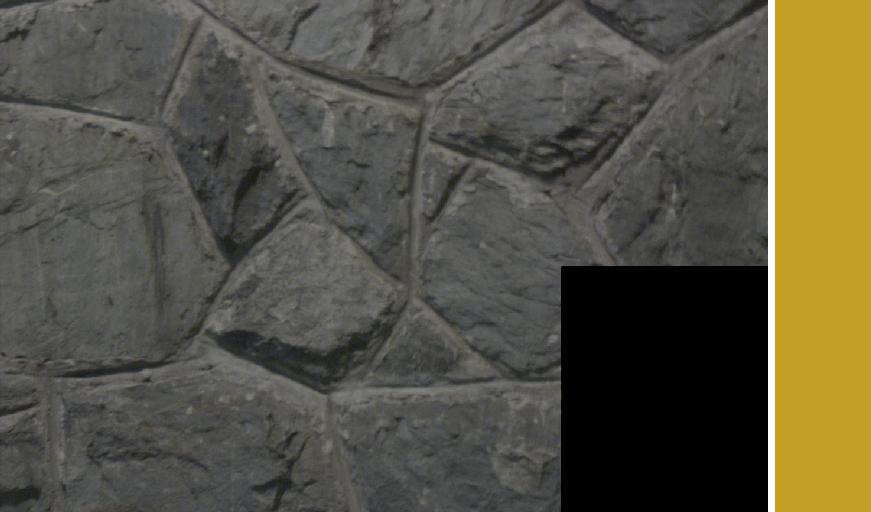}
    \vspace{-1.5em}
    \caption*{Ground Truth}
\end{subfigure}


\begin{subfigure}{\xxwidth\textwidth}
    \centering
    \includegraphics[width=0.88\columnwidth]{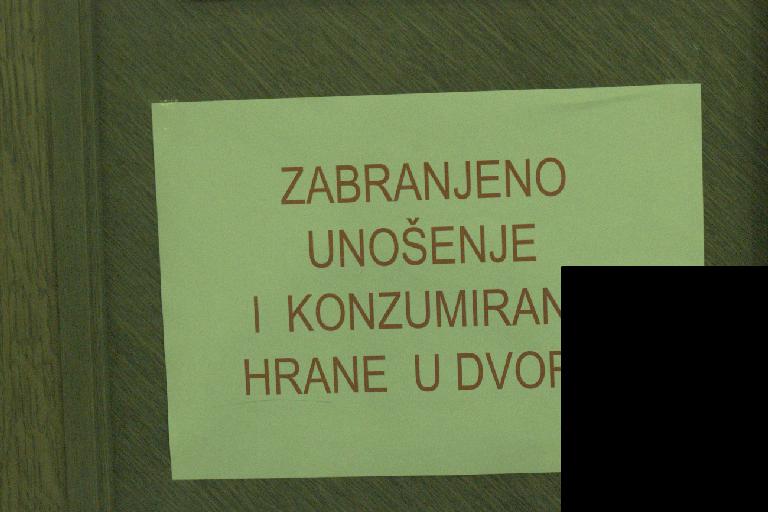}
    \vspace{-0.4em}
    \caption*{Input}
\end{subfigure}
\begin{subfigure}{\xxwidth\textwidth}
    \centering
    \includegraphics[width=\columnwidth]{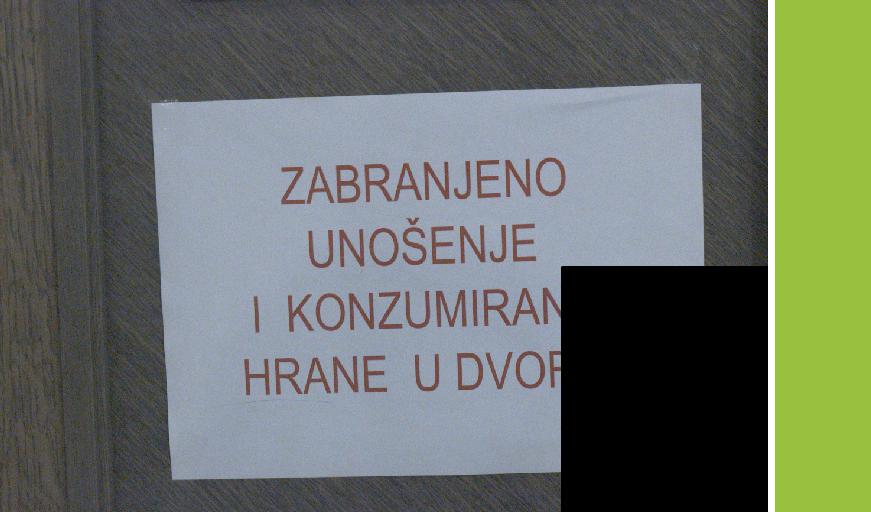}
    \vspace{-1.5em}
    \caption*{FC4~($3.58^{\circ}$)}
\end{subfigure}
\begin{subfigure}{\xxwidth\textwidth}
    \centering
    \includegraphics[width=\columnwidth]{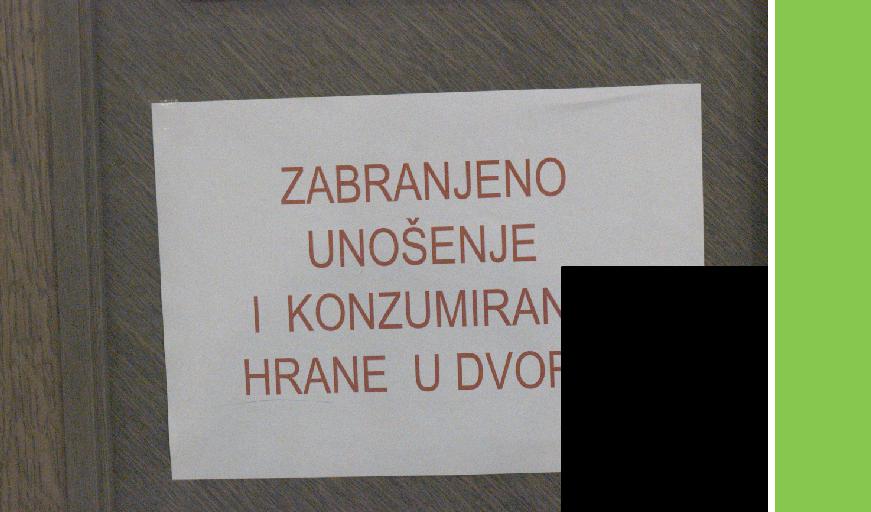}
    \vspace{-1.5em}
    \caption*{C4~($1.92^{\circ}$)}
\end{subfigure}
\begin{subfigure}{\xxwidth\textwidth}
    \centering
    \includegraphics[width=\columnwidth]{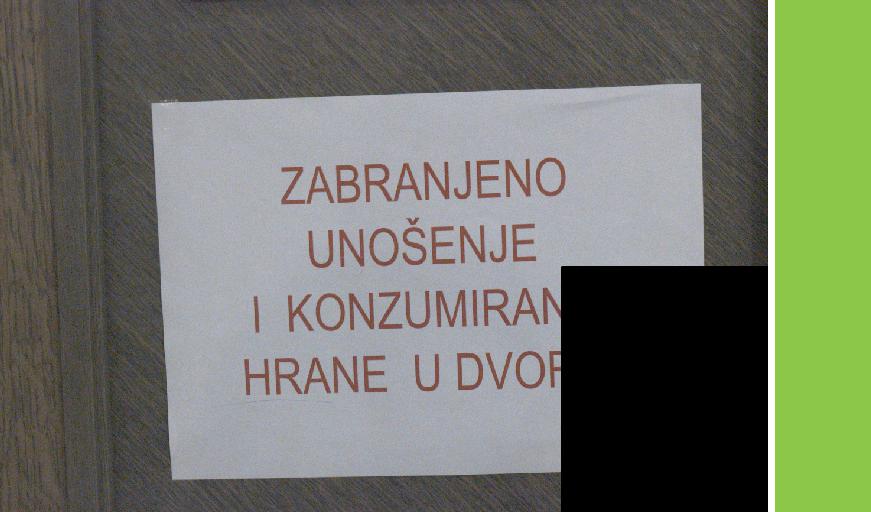}
    \vspace{-1.5em}
    \caption*{Ours~($\mathbf{0.35^{\circ}}$)}
\end{subfigure}
\begin{subfigure}{\xxwidth\textwidth}
    \centering
    \includegraphics[width=\columnwidth]{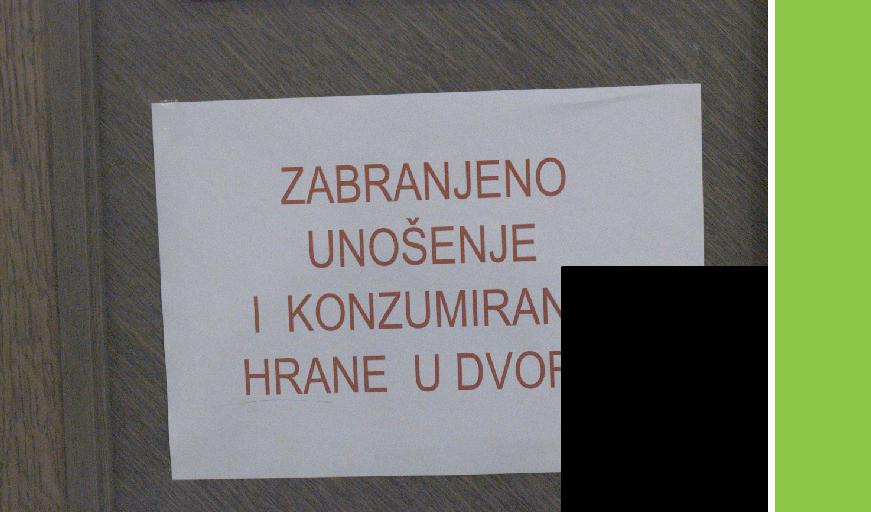}
    \vspace{-1.5em}
    \caption*{Ground Truth}
\end{subfigure}


\begin{subfigure}{\xxwidth\textwidth}
    \centering
    \includegraphics[width=0.88\columnwidth]{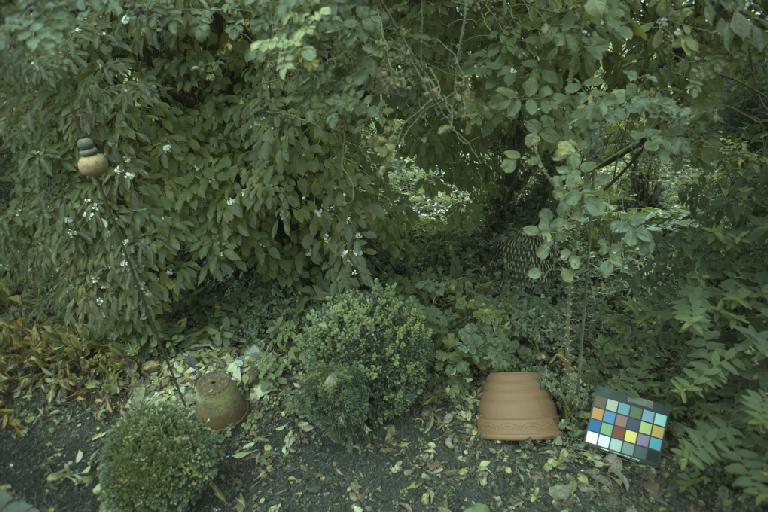}
    \vspace{-0.5em}
    \caption*{Input}
\end{subfigure}
\begin{subfigure}{\xxwidth\textwidth}
    \centering
    \includegraphics[width=\columnwidth]{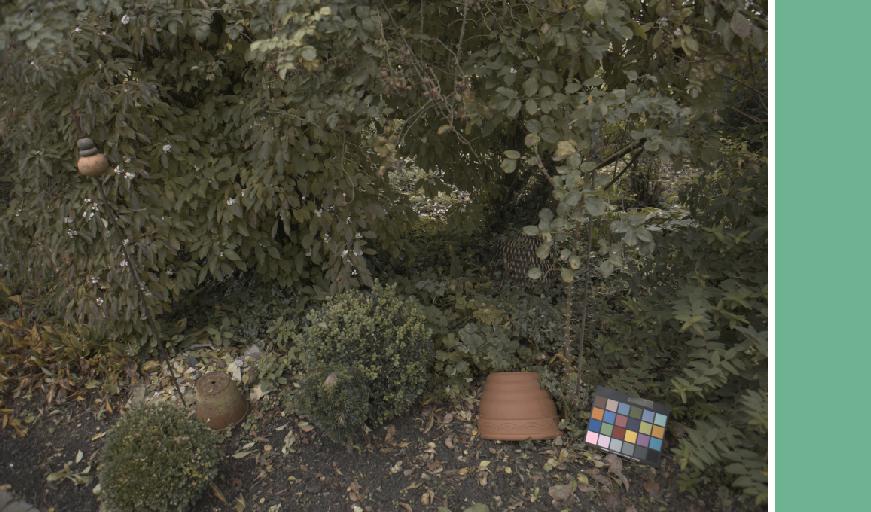}
    \vspace{-1.5em}
    \caption*{FC4~($2.77^{\circ}$)}
\end{subfigure}
\begin{subfigure}{\xxwidth\textwidth}
    \centering
    \includegraphics[width=\columnwidth]{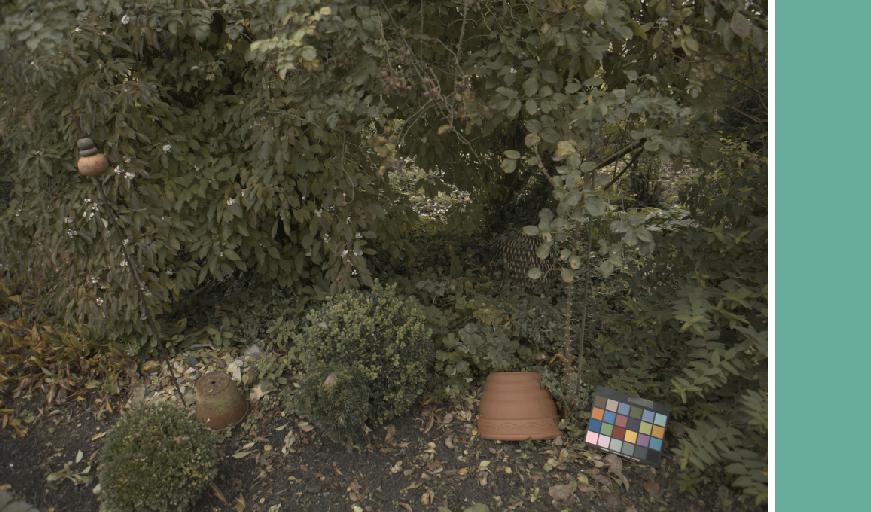}
    \vspace{-1.5em}
    \caption*{C4~($1.32^{\circ}$)}
\end{subfigure}
\begin{subfigure}{\xxwidth\textwidth}
    \centering
    \includegraphics[width=\columnwidth]{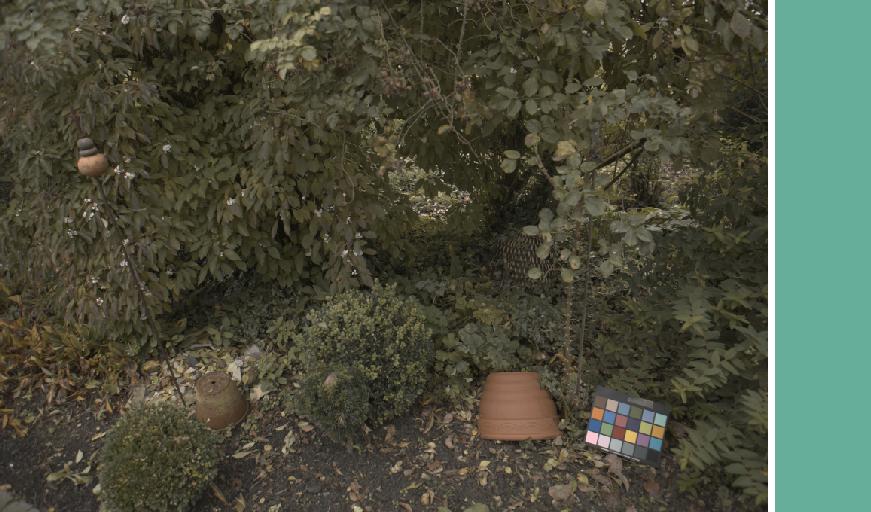}
    \vspace{-1.5em}
    \caption*{Ours~($\mathbf{0.67^{\circ}}$)}
\end{subfigure}
\begin{subfigure}{\xxwidth\textwidth}
    \centering
    \includegraphics[width=\columnwidth]{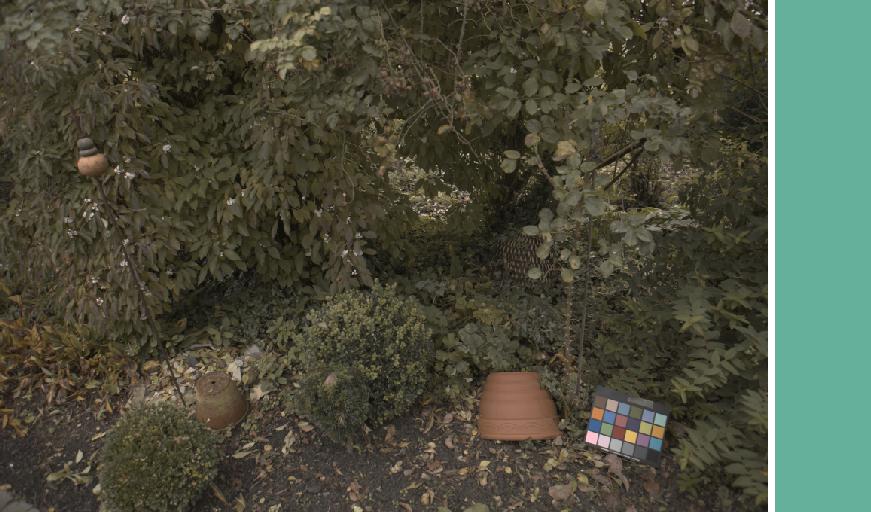}
    \vspace{-1.5em}
    \caption*{Ground Truth}
\end{subfigure}


\begin{subfigure}{\xxwidth\textwidth}
    \centering
    \includegraphics[width=0.88\columnwidth]{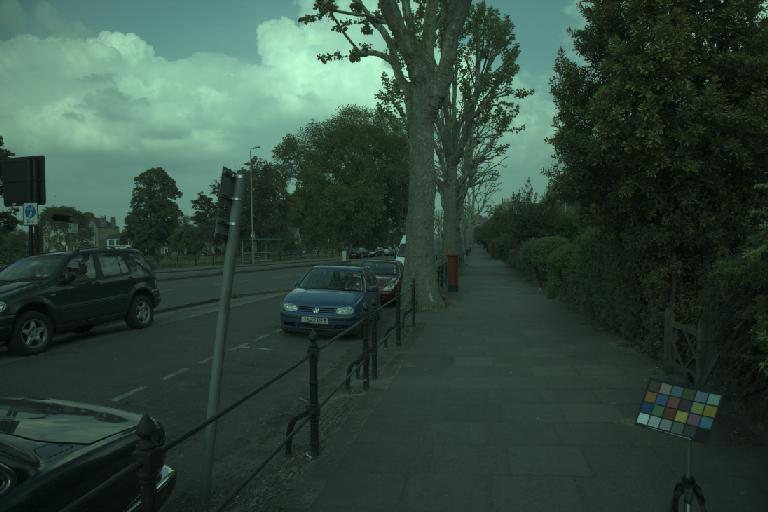}
    \vspace{-0.5em}
    \caption*{Input}
\end{subfigure}
\begin{subfigure}{\xxwidth\textwidth}
    \centering
    \includegraphics[width=\columnwidth]{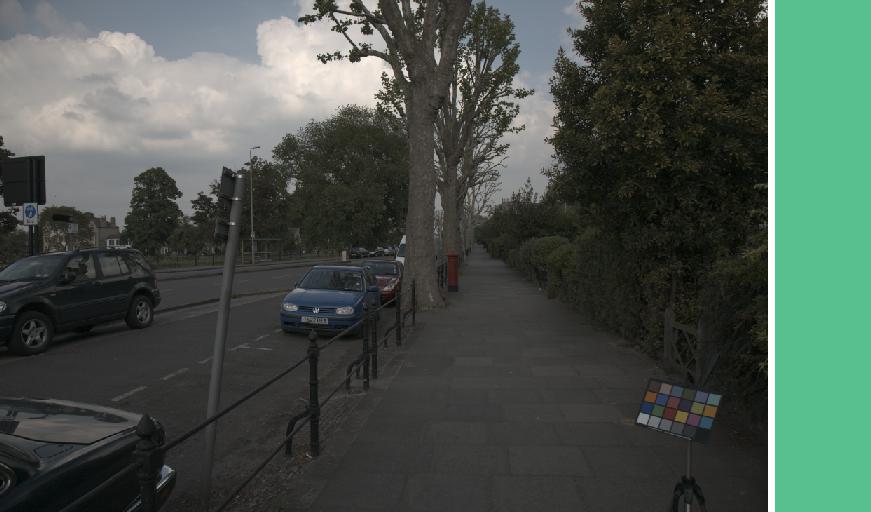}
    \vspace{-1.5em}
    \caption*{FC4~($2.17^{\circ}$)}
\end{subfigure}
\begin{subfigure}{\xxwidth\textwidth}
    \centering
    \includegraphics[width=\columnwidth]{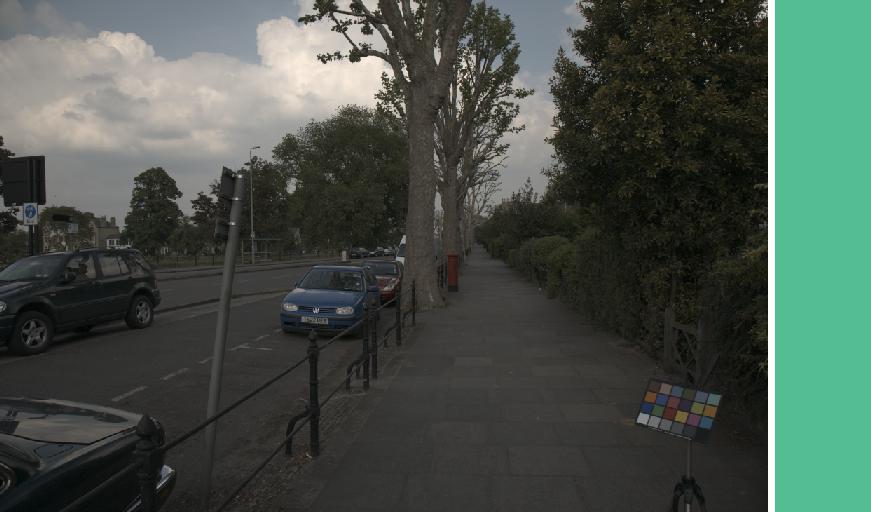}
    \vspace{-1.5em}
    \caption*{C4~($0.91^{\circ}$)}
\end{subfigure}
\begin{subfigure}{\xxwidth\textwidth}
    \centering
    \includegraphics[width=\columnwidth]{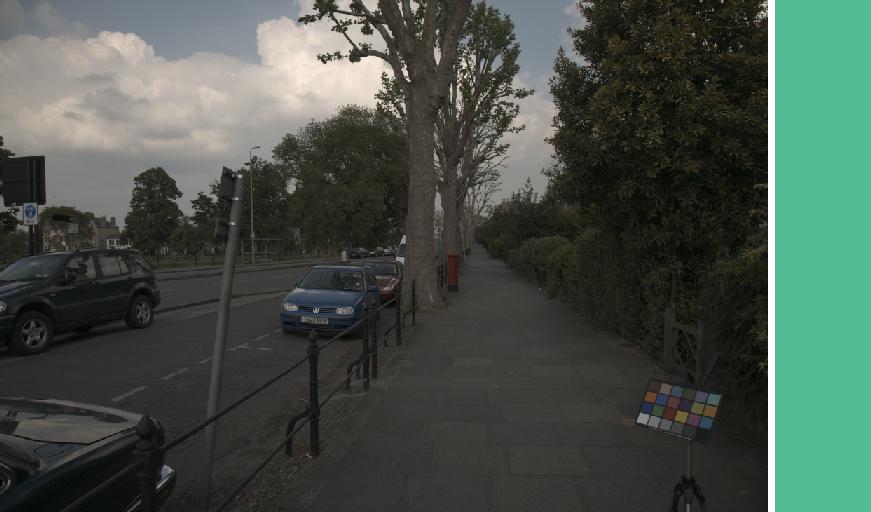}
    \vspace{-1.5em}
    \caption*{Ours~($\mathbf{0.42^{\circ}}$)}
\end{subfigure}
\begin{subfigure}{\xxwidth\textwidth}
    \centering
    \includegraphics[width=\columnwidth]{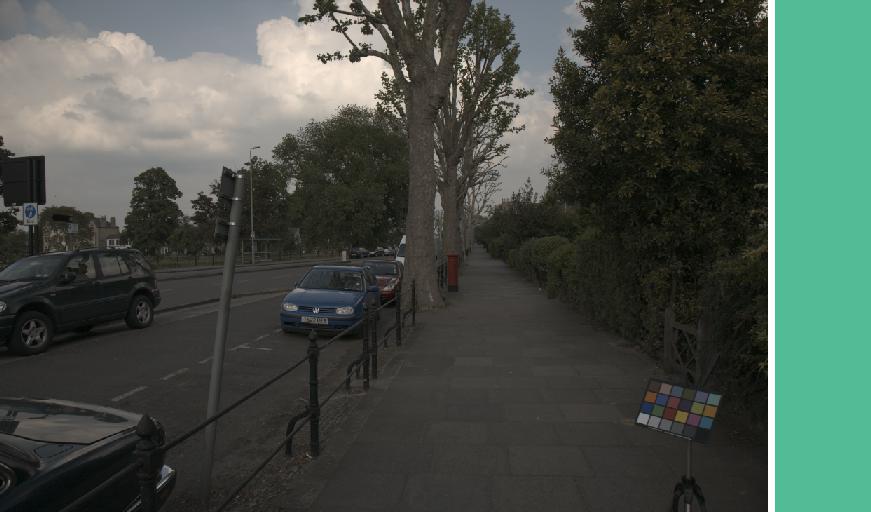}
    \vspace{-1.5em}
    \caption*{Ground Truth}
\end{subfigure}


\caption{Comparisons among the proposed method, C4~\cite{yu2020c4} and FC4~\cite{hu2017fc4}. For each method, we visualize the corrected image on the left of each predicted illuminant, and report the angular error compared with the corresponding ground truth below each image. The upper two images are taken from Cube-Challenge while the other two are from Gehler-Shi. Note that the calibration objects are masked out when training and testing.}
\label{fig:visual}
\end{figure*}

\begin{table}[t]
    \small 
    \centering
    \caption{Comparison with the state-of-the-art methods on the Cube-Challenge dataset. The \colorbox{lightgray!30}{shadowed} methods are for single-sensor evaluation while the others are for  cross-sensor evaluation.}

    \begin{tabular}{l|ccccc}
    \toprule
    Method &  Mean & Med. & Tri. & Best & Worst \\
     &   &  &  & 25\% & 25\% \\\hline
    Grey World~\cite{gray-world} & 4.44 & 3.50 & - & 0.77 & 9.64 \\
    1st Gray-Edge~\cite{gray-edge}& 3.51 & 2.30 & - & 0.56 & 8.53 \\
    V Vuk \etal~\cite{cubechallenge}& 6.00 & 1.96 & 2.25 & 0.99 & 18.81 \\
    SIIE~\cite{siie}&  2.10 & 1.23 & - & 0.47 & 5.38 \\
    Multi-Hyp.~\cite{multi-hyp} & 1.99 & \textbf{1.06} & \textbf{1.14} & \textbf{0.35} & 5.35 \\
    Ours~(cross-sensor) & \textbf{1.75} & 1.26 & 1.32 & 0.42 & \textbf{4.04} \\ \hline
    \rowcolor{lightgray!30} Y Qian \etal~\cite{cubechallenge}& 2.21 & 1.32 & 1.41 & 0.43 & 5.65 \\
    \rowcolor{lightgray!30} K Chen \etal~\cite{cubechallenge}& 1.84 & 1.27 & 1.32 & 0.39 & 4.41 \\
    \rowcolor{lightgray!30} FFCC~(model J)~\cite{ffcc}& 2.10 & 1.23 & 1.34 & 0.47 & 5.38 \\
    \rowcolor{lightgray!30} A Savchik \etal~\cite{cubechallenge}& 2.05 & 1.20 & 1.30 & 0.40 & 5.24 \\
    \rowcolor{lightgray!30} WB-sRGB~\cite{afifi2019color}& 1.83 & 1.15 & - & 0.35 & 4.60 \\
    \rowcolor{lightgray!30} FC4~\cite{hu2017fc4}  &  1.75 & 1.00 & 1.13 & 0.37 & 4.45\\
    \rowcolor{lightgray!30} C4~\cite{yu2020c4}  & 1.76 & 1.11 & 1.25 &  0.37 &   4.54  \\

    \rowcolor{lightgray!30} Ours~(single-sensor) & 1.57 & \textbf{0.91} & 1.06 & 0.31 & 3.92  \\
    \rowcolor{lightgray!30} Ours~(fine-tuning) & \textbf{1.47} & 0.95 & \textbf{1.02} & \textbf{0.30} & \textbf{3.60}  \\
    \bottomrule
    \end{tabular}
    \label{tab:cubec}
\end{table}

\subsection{Comparison with State-of-the-Art Methods}
\label{sec:exp_compare}
We compare our method with a number of representative algorithms, including widely-used statistics-based~\cite{white-patch,gray-world,gray-edge,shades-of-gray}, early machine learning-based~\cite{SVR,gehler2008bayesian} and recent deep learning-based~\cite{bianco2015color,cm,ccc,ffcc,hu2017fc4,yu2020c4,siie,end2end}.

The numeric results on the widely-used  Gehler-Shi, NUS8 and Cube-Challenge datasets are reported in Tables~\ref{tab:compare} and~\ref{tab:cubec}. All our experiments are trained on the proposed network. In the cross-sensor setting, we train the network on COCO~\cite{coco} and INTEL-TAU~\cite{laakom2019intel} with the proposed  self-supervised cross-sensor training method and evaluate its performance. In the single-sensor experiments, without UIP and SAF, we train our network from scratch using the sensor-aware illuminant enhancement~(SIE) described in Section~\ref{sec32}.

As for the cross-sensor evaluation, although our model is fully trained with the synthesized illuminants, it still obtains significantly better performance than other cross-sensor methods by a large margin. On the carefully collected single-sensor evaluation datasets~(NUS8 and Cube-Challenge), our cross-sensor model even performs better than most of single-sensor models owing to the proposed UIP and SAF. Also, we argue that compared with previous cross-sensor methods, training with UIP and SAF can gain much more \textit{robust} results in terms of the metric of Worst 25\%. 

On the other hand, the proposed network structure is also the key to achieve the state-of-the-art performance. We evaluate the performance of the proposed network trained \textit{from scratch} and it shows a much better performance than other methods in Tables~\ref{tab:compare} and~\ref{tab:cubec}~(see the single-sensor comparison). Finally, as shown in the last row of Table~\ref{tab:cubec}, by fine-tuning the cross-sensor model on the specific sensor, the performance can be further boosted. 


In addition to the numerical comparisons, we also show some visual comparisons on Gehler-Shi and Cube-Challenge with FC4 and C4. To get the visual results, we directly use the trained models from the Github page\footnote{\url{https://github.com/yhlscut/C4}.} on the Gehler-Shi dataset, and train them from scratch on Cube-Challenge with the default settings. As shown in Figure~\ref{fig:visual}, our method shows a significant advantage over the two baselines. We also find our model is very good at handling the situation when the scenes contain few objects~(top two samples in Figure~\ref{fig:visual}), which might be because the proposed sensor-aware illuminant enhancement is based on the von Kries assumption and these scenes are more consistent with this assumption.  

\begin{table*}[!t]
    \centering
    \caption{The mean of angular errors on \textit{cross-sensor} evaluation which directly evaluates pre-trained methods on various sensors without single-sensor fine-tuning. Both the methods Naive and SAF use the INTEL-TAU dataset~\cite{laakom2019intel} for training, while UIP uses COCO~\cite{coco}. Notice that the sensors from the training sets are different from those in the testing sets (Gehler-Shi, NUS8 and Cube-Challenge). The {\color{blue}blue} and {\color{red}red} numbers are the relative {\color{blue}improvement} and {\color{red}decline}, compared with the model trained by Naive method. All the methods are based on the proposed network.} 
    \resizebox{\textwidth}{!}{%
    \begin{tabular}{l|llllllll|l|l}
    \toprule
    Datasets & \multicolumn{8}{c|}{NUS8} & Cube-Challenge & \multicolumn{1}{c}{Gehler-Shi} \\ \hline
    Method & \textit{Canon 1D} & \textit{Canon 600D} & \textit{Fuji} & \textit{Nikon 52} & \textit{Olympus} & \textit{Panasonic} & \textit{Samsung} & \textit{Sony} & \textit{Canon 550D} & \textit{Canon 5D\&1D} \\ \hline
     Naive &3.07 &2.61 &3.66 &3.45 &3.25 &2.41 &2.92 &3.49 &2.76 & 3.71\\
     SAF &2.09~({\color{blue}{\footnotesize0.98}})
     &2.07~({\color{blue}{\footnotesize0.54}})
     &2.15~({\color{blue}{\footnotesize1.51}})
     &2.14~({\color{blue}{\footnotesize1.31}})
     &\textbf{1.88}~({\color{blue}{\footnotesize1.37}})
     &2.13~({\color{blue}{\footnotesize0.28}})
     &2.05~({\color{blue}{\footnotesize0.87}})
     &1.95~({\color{blue}{\footnotesize1.54}})
     &1.91~({\color{blue}{\footnotesize0.85}})
     &2.58~({\color{blue}{\footnotesize1.13}})\\
     UIP &6.82~({\color{red}{\footnotesize3.75}})
     &9.24~({\color{red}{\footnotesize6.63}})
     &8.12~({\color{red}{\footnotesize4.46}})
     &9.48~({\color{red}{\footnotesize6.03}})
     &9.07~({\color{red}{\footnotesize5.82}})
     &10.44~({\color{red}{\footnotesize8.03}})
     &11.59~({\color{red}{\footnotesize8.67}})
     &7.51~({\color{red}{\footnotesize4.02}})
     &5.53~({\color{red}{\footnotesize2.77}})
     &7.27~({\color{red}{\footnotesize3.56}})\\
    UIP\&SAF &\textbf{1.99}~({\color{blue}{\footnotesize1.08}})
    &\textbf{1.80}~({\color{blue}{\footnotesize0.81}})
    &\textbf{2.12}~({\color{blue}{\footnotesize1.54}})
    &\textbf{2.11}~({\color{blue}{\footnotesize1.34}})
    &1.93~({\color{blue}{\footnotesize1.32}})
    &\textbf{1.94}~({\color{blue}{\footnotesize0.47}})
    &\textbf{2.04}~({\color{blue}{\footnotesize0.88}})
    &\textbf{1.90}~({\color{blue}{\footnotesize1.59}})
    &\textbf{1.75}~({\color{blue}{\footnotesize1.01}})
    & \textbf{2.46}~({\color{blue}{\footnotesize1.25}}) \\
    \bottomrule
    \end{tabular}}
    \label{tab:xcc}
\end{table*}

\begin{table*}[!t]
    \centering
    \small
    \caption{The mean of angular errors on \textit{single-sensor} fine-tuning evaluation using different pre-trained cross-sensor methods on 8 NUS8 subsets and Cube-Challenge~(\textit{Canon 550D}). The {\color{blue}blue} and {\color{red}red} numbers are the relative {\color{blue}improvement} and {\color{red}decline}, compared with the model trained from scratch. All the methods are based on the proposed network.}
    \begin{tabular}{l|lllllllll}
    \toprule
    Method & \textit{Canon 1D} & \textit{Canon 600D} & \textit{Fuji} & \textit{Nikon 52} & \textit{Olympus} & \textit{Panasonic} & \textit{Samsung} & \textit{Sony} & \textit{Canon 550D} \\ \hline
    Scratch &1.84 &1.67 &1.86 &1.85 &\textbf{1.43} &1.77 &1.78 &1.59 &1.57\\
    Naive &1.77~({\color{blue}{\footnotesize0.07}}) &1.67~({\color{black}{\footnotesize0.0}}) &1.79~({\color{blue}{\footnotesize0.07}}) & 1.77~({\color{blue}{\footnotesize0.08}}) & 1.57~({\color{red}{\footnotesize0.14}}) &1.77~({\color{black}{\footnotesize0.0}}) &1.78~({\color{black}{\footnotesize0.0}}) &1.64~({\color{red}{\footnotesize0.05}}) &1.57~({\color{black}{\footnotesize0.0}})\\
    SAF &\textbf{1.74}~({\color{blue}{\footnotesize0.10}}) &1.62~({\color{blue}{\footnotesize0.05}}) &1.71~({\color{blue}{\footnotesize0.15}}) &1.84~({\color{blue}{\footnotesize0.01}}) &1.53~({\color{red}{\footnotesize0.10}}) &1.71~({\color{blue}{\footnotesize0.06}}) & 1.79~({\color{red}{\footnotesize0.01}}) &1.56~({\color{blue}{\footnotesize0.03}}) &1.51~({\color{blue}{\footnotesize0.06}})\\
    UIP &1.77~({\color{blue}{\footnotesize0.07}}) &\textbf{1.52}~({\color{blue}{\footnotesize0.15}}) &\textbf{1.68}~({\color{blue}{\footnotesize0.18}}) &1.82~({\color{blue}{\footnotesize0.03}}) &1.46~({\color{red}{\footnotesize0.03}}) &\textbf{1.60}~({\color{blue}{\footnotesize0.17}}) &1.77~({\color{blue}{\footnotesize0.01}}) &\textbf{1.54}~({\color{blue}{\footnotesize0.05}}) &1.49~({\color{blue}{\footnotesize0.08}})\\
    UIP\&SAF &1.78~({\color{blue}{\footnotesize0.06}}) & 1.58~({\color{blue}{\footnotesize0.09}}) &1.70~({\color{blue}{\footnotesize0.16}}) &\textbf{1.74}~({\color{blue}{\footnotesize0.11}}) &1.48~({\color{red}{\footnotesize0.05}}) &1.61~({\color{blue}{\footnotesize0.16}}) &\textbf{1.72}~({\color{blue}{\footnotesize0.06}}) &1.56~({\color{blue}{\footnotesize0.03}}) &\textbf{1.47}~({\color{blue}{\footnotesize0.10}})\\
    \bottomrule
    \end{tabular}
    \label{tab:finetune}
\end{table*}

\begin{figure*}[t!]
    \centering
    \begin{subfigure}[t]{0.8\columnwidth}
        \centering
        \includegraphics[width=\columnwidth]{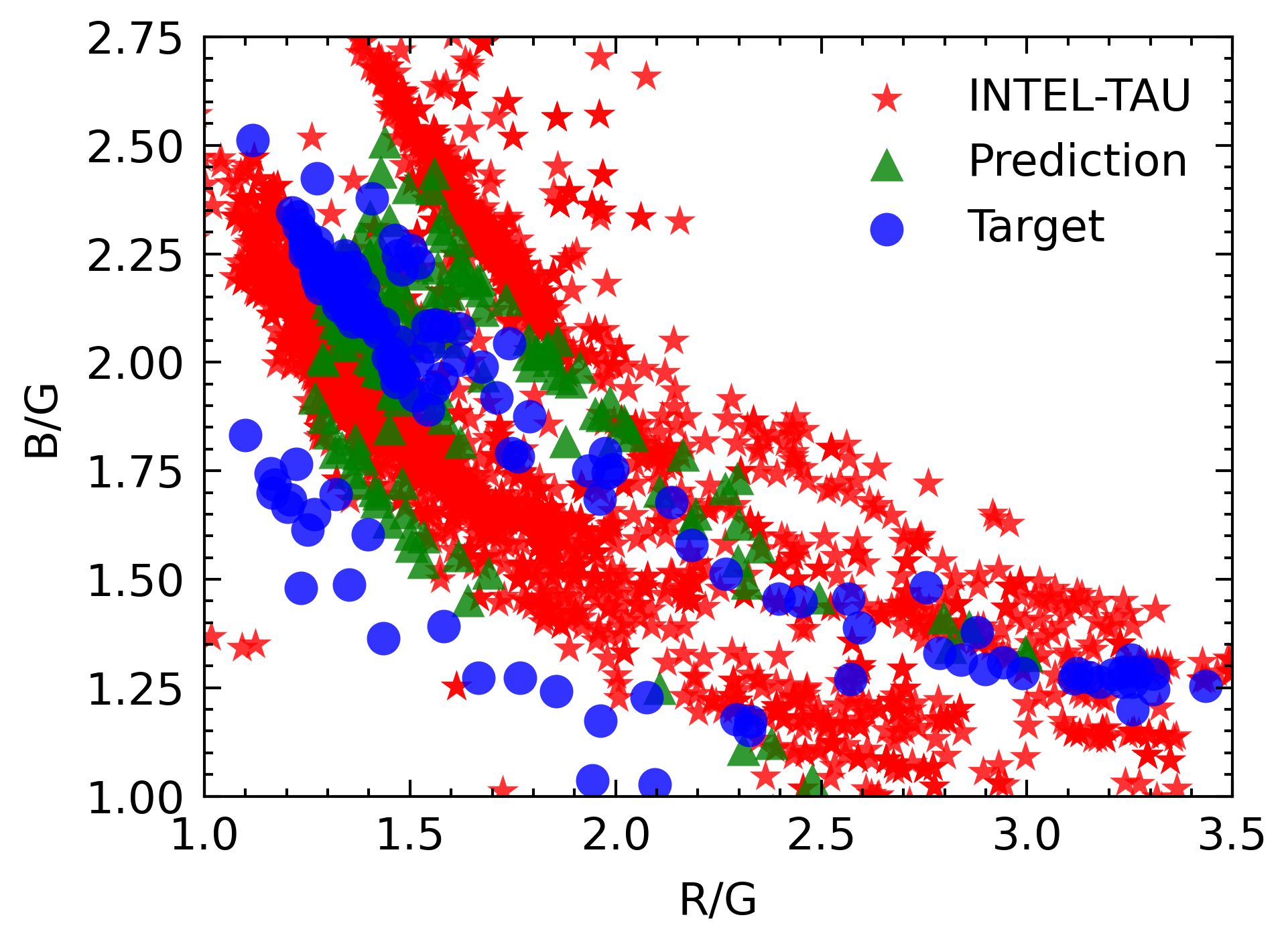}
        \vspace{-1em}
        \caption{Naive}
        \label{fig:cda}
    \end{subfigure}%
    \begin{subfigure}[t]{0.8\columnwidth}
        \centering
        \includegraphics[width=\columnwidth]{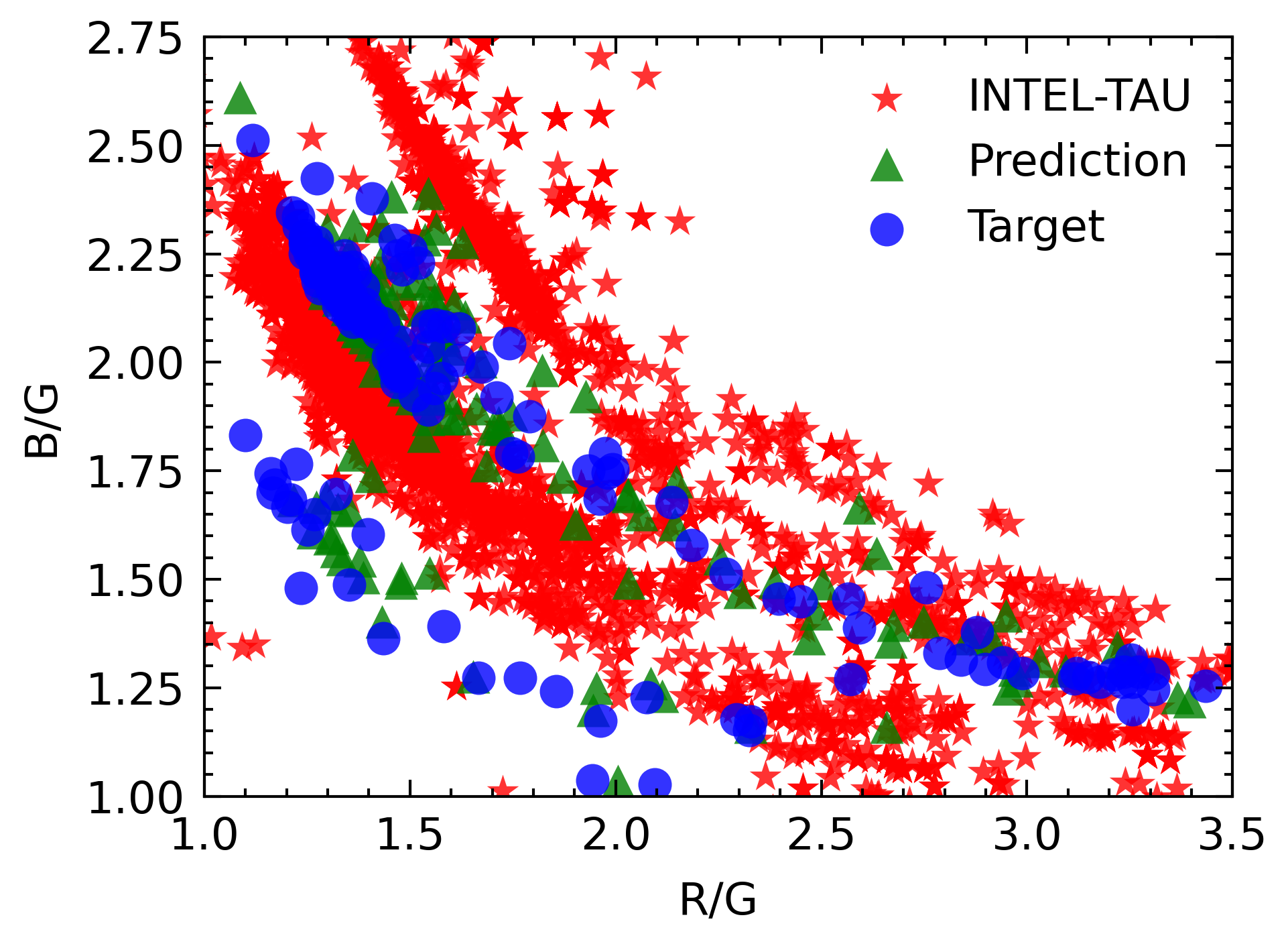}
        \vspace{-1em}
        \caption{SAF}
        \label{fig:cdb}
    \end{subfigure}
    \caption{The effect of SAF in cross-sensor training. (a)~The distribution of illuminants in the $[\frac{G}{B},\frac{G}{R}]$ color space, which are predicted by our network trained on INTEL-TAU and tested on Gehler-Shi~\cite{gehler2008bayesian,shi2000re}. (b)~The proposed SAF trains the network with randomly generated illuminants, which successfully fits the evaluation sensor's domain. }
    \label{fig:cd}
\end{figure*}

As for the model size, the proposed network uses only 17\% of the parameters of C4 and 47\% of FC4, which further shows the advantage and the potential of our method applying on embedded devices, such as mobile phones and digital cameras.

\subsection{Evaluation of Self-Supervised Cross-Sensor Training}
\label{sec:exp_cross}
Firstly, we evaluate the effect of the proposed unprocessing sRGB Images for Pre-Training~(UIP) and Sensor Alignment Fine-Tuning~(SAF) on three unseen datasets~(Gehler-Shi, NUS8 and Cube-Challenge) as shown in Table~\ref{tab:xcc}. We can see that although SAF without UIP trains the network with fully synthesized random illuminants, it obtains better results than training with only real illuminants~(Naive). We show an example in Figure~\ref{fig:cd} to compare the performance of the network using Naive and SAF. The network using Naive predicts a wrong color distribution for the target sensor~(a), while the network using SAF can fit the target distribution well~(b). These results show the advantage of training on the aligned domain using random illuminants.

   

As for UIP, since the images $I_{uip}$ (unprocessed from sRGB images) are still different from real RAW images $I_{raw}$, the performance of the network using only UIP is not good enough when evaluated on the RAW images in the testing datasets. However, the advantage of UIP is that sRGB images are abundant and the network can learn the knowledge of different scenes and illuminants. We further fine-tune the network with SAF~(UIP\&SAF), it obtains the best results in Table~\ref{tab:xcc}, and outperforms other state-of-the-art cross-sensor methods as shown in Tables~\ref{tab:compare} and \ref{tab:cubec}.

We also evaluate the transferable ability of the different pre-trained cross-sensor models for single-sensor fine-tuning.
As shown in Table~\ref{tab:finetune}, the proposed unprocessing sRGB images for pre-training~(UIP), sensor alignment fine-tuning~(SAF) and UIP\&SAF training strategies can improve the performance of the single-sensor model trained from scratch. Note that when we just pre-train our network with images from multiple sensors~(Naive), the performance cannot be improved in most cases~( \textit{Canon 600D, Olympus, Panasonic, Samsung, Sony, Canon 550D} in the second row of Table~\ref{tab:finetune}).  Interestingly, although the UIP model cannot obtain satisfactory results when we evaluate it directly on specific sensors~(as shown in Table~\ref{tab:xcc}), this model has a strong transferable ability when fine-tuning it on a single sensor~(as shown in Table~\ref{tab:finetune}). It might be because the large-scale semantic information learned by UIP benefits accurate color constancy. To summarize, the UIP\&SAF method can improve the performance by a large range in most cases. 


\begin{table}[t]
    \centering
     \caption{Effect of the percentage of different training combinations. The best results are bold and the second best are underlined.}
    \resizebox{\columnwidth}{!}{%
    
    \begin{tabular}{c|c|c|c|c}
    \toprule
    Dataset & Random & No Aug. & Reshuffle & Mean  \\ \hline
    \textit{Sony}    &  100\% & 0  &  0   & 1.97 \\
    \textit{Sony}    &  0  & 100\% &  0   & 1.92 \\ 
     \textit{Sony}   &  0  & 0     & 100\% & 1.80 \\
    \textit{Sony}    &  0  & 50\%  & 50\%  & \textbf{1.76} \\
    \textit{Sony}    &  33\% & 33\% & 34\% & \underline{1.79} \\ 
    \hline
    \textit{Nikon 52}   &  100\% & 0  &  0   & 2.30 \\
    \textit{Nikon 52}   &  0  & 100\% &  0   & 2.26 \\ 
    \textit{Nikon 52}   &  0  & 0     & 100\%  & \textbf{2.05} \\
    \textit{Nikon 52}    &  0  & 50\%  & 50\%  & 2.13 \\
    \textit{Nikon 52}    &  33\% & 33\% & 34\% & \underline{2.07} \\
    \bottomrule
    \end{tabular}
}
\label{tab:precentage}
\end{table}

\begin{figure}[t]
    \centering
    \includegraphics[width=0.8\columnwidth]{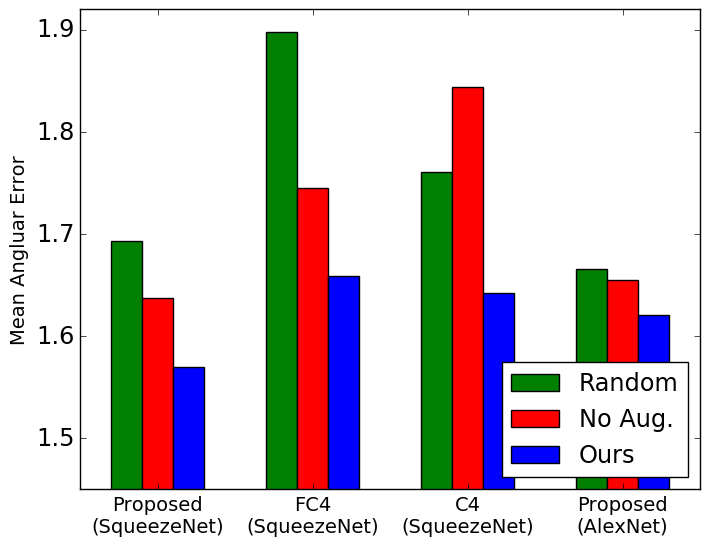}
    \caption{Performances of different relighting methods on various models.}
    \label{fig:relighting_methods}
\end{figure}

\begin{figure}[t]
    \centering
    \includegraphics[width=\columnwidth]{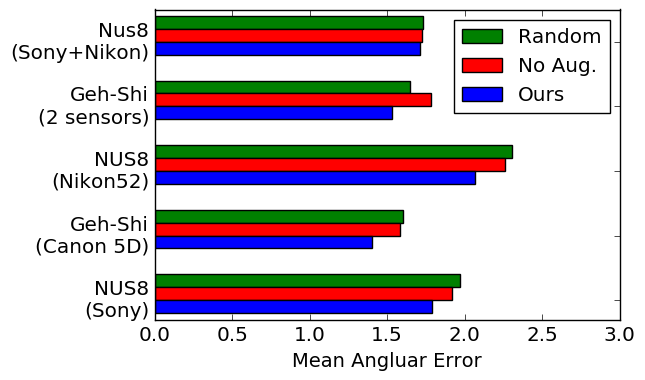}
    \caption{Performances of different relighting methods on various datasets.}
    \label{fig:ab_more_relight}
\end{figure}


\subsection{Evaluation of Sensor-Aware Illuminant Enhancement for Single-Sensor Training}
\label{sec:exp_single}

In this section, we discuss the effect of the sensor-aware illuminant enhancement when training from \textit{scratch} for single-sensor training. 

\textbf{Percentages on Different Training Combinations.}
As discussed in Section~\ref{sec32}, we alternatively use three different types of relighting methods on the fly to train our network. Here, we evaluate the percentages of the training combinations among the three methods on the \textit{Nikon 52} and \textit{Sony} subsets of NUS8 using FC4~\cite{hu2017fc4}, as shown in Table~\ref{tab:precentage}. It is clear that using the sensor-aware label reshuffle~(Reshuffle) solely performs better than no-relighting~(No Aug.) and random relighting~(Random~\cite{hu2017fc4}). Moreover, we find that the network may get further improvement and robustness when there are multiple training combinations. Thus, in all the experiments, we equally choose samples from these three training combinations, which is a hybrid training strategy for single-sensor training.

\textbf{Scene Relighting Methods on Different Models.}
We train different networks, including SqueezeNet-FC4~\cite{hu2017fc4}, SqueezeNet-C4~\cite{yu2020c4} and the proposed network~(with either SqueezeNet or AlexNet as the backbone), using different data augmentation methods on Cube-Challenge as shown in Figure~\ref{fig:relighting_methods}. The proposed sensor-aware illuminant enhancemant can significantly improve the performances of these networks, especially for FC4. We also note that compared with no-relighting, the random relighting also benefits C4 since C4 is very heavy~(its parameters are 6 times more than ours) and needs more training data to avoid overfitting.


\textbf{Effect on Different Datasets.}
We also evaluate different scene relighting methods as data augmentation for individual and multiple sensors using FC4~\cite{hu2017fc4}. For single camera evaluation, we train FC4 and evaluate its performance on the \textit{Sony} and \textit{Nikon52} subsets of NUS8~\cite{nus8} and the \textit{Canon 5D} part of the Gehler-Shi dataset. As for multiple camera learning, the evaluations are performed on the full Gehler-Shi dataset and on the combination of the \textit{Sony} and \textit{Nikon52} subsets. As shown in Figure~\ref{fig:ab_more_relight}, The proposed hybrid training strategy obtains much better results than the previous random relighting~(Random) and no-relighting~(No~Aug.) methods on both the single and multiple camera settings. Interestingly, we find random relighting~\cite{hu2017fc4} 
only performs better on the Gehler-Shi dataset than no-relighting. It might be because Gehler-Shi contains extremely imbalanced sensor data distribution~(482~(\textit{Canon 5D})~:~86~(\textit{Canon 1D})) and the small subset~(86) provides little useful camera property guidance.
Thus, we argue that although random relighting has been widely-used in recent methods~\cite{hu2017fc4,mdlcc,yu2020c4}, it might \textit{hurt} the performance in most cases.

\begin{figure*}[t]
\centering
\begin{subfigure}{\xxwidth\textwidth}
    \centering
    \includegraphics[width=0.88\columnwidth]{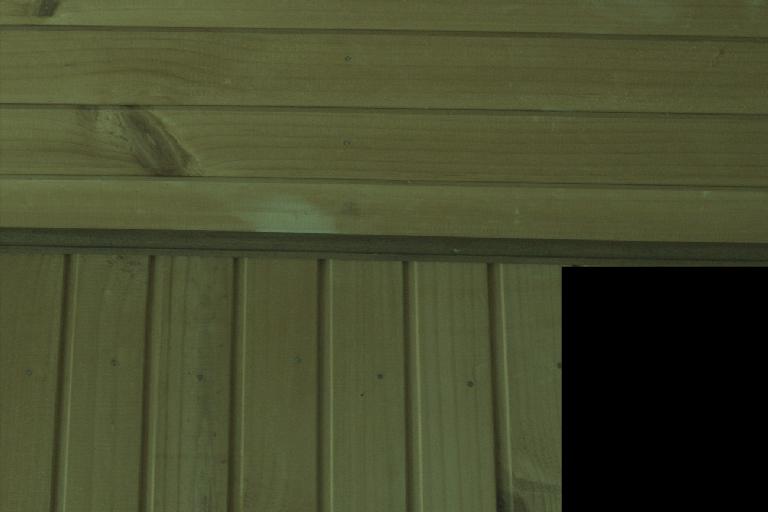}
    \vspace{-0.4em}
    \caption*{Input}
\end{subfigure}
\begin{subfigure}{\xxwidth\textwidth}
    \centering
    \includegraphics[width=\columnwidth]{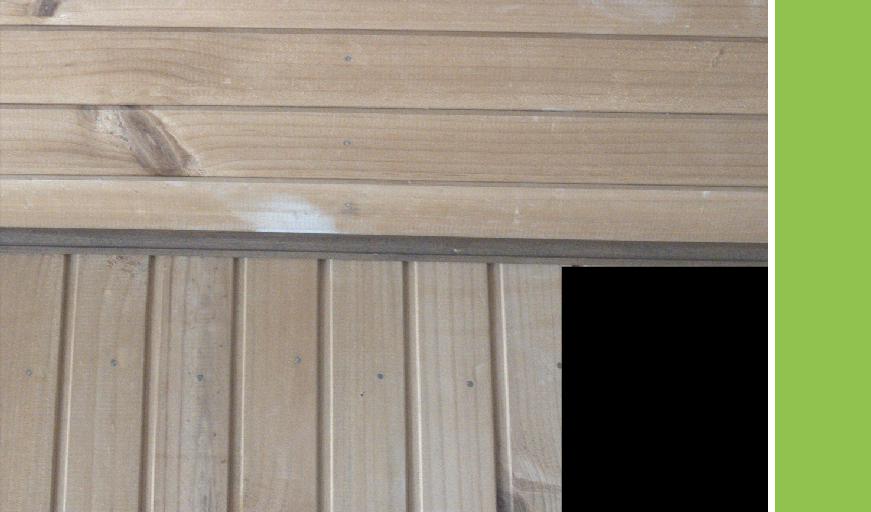}
    \vspace{-1.5em}
    \caption*{Stage1~($5.75^{\circ}$)}
\end{subfigure}
\begin{subfigure}{\xxwidth\textwidth}
    \centering
    \includegraphics[width=\columnwidth]{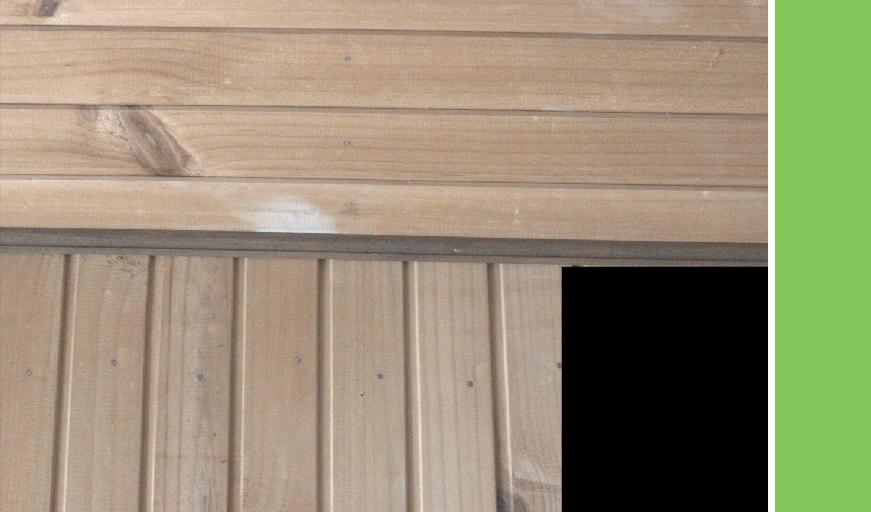}
    \vspace{-1.5em}
    \caption*{Stage2~($2.20^{\circ}$)}
\end{subfigure}
\begin{subfigure}{\xxwidth\textwidth}
    \centering
    \includegraphics[width=\columnwidth]{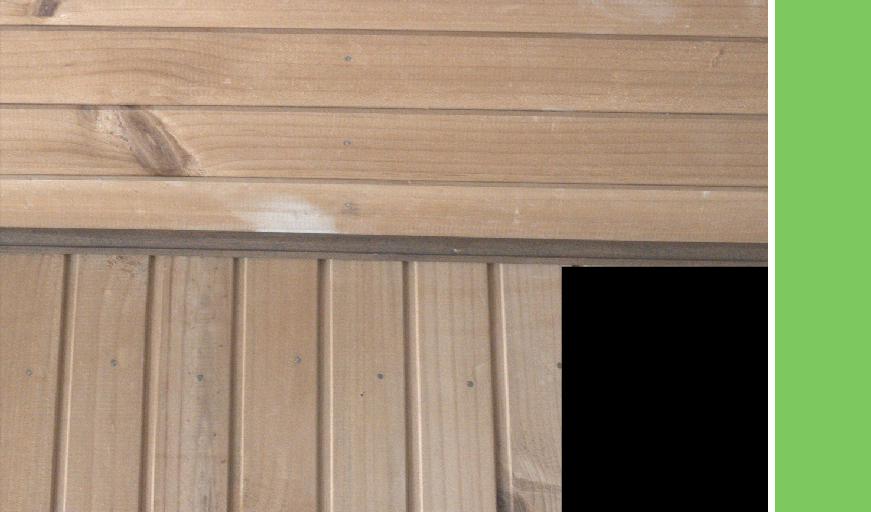}
    \vspace{-1.5em}
    \caption*{Stage3~($\mathbf{1.45^{\circ}}$)}
\end{subfigure}
\begin{subfigure}{\xxwidth\textwidth}
    \centering
    \includegraphics[width=\columnwidth]{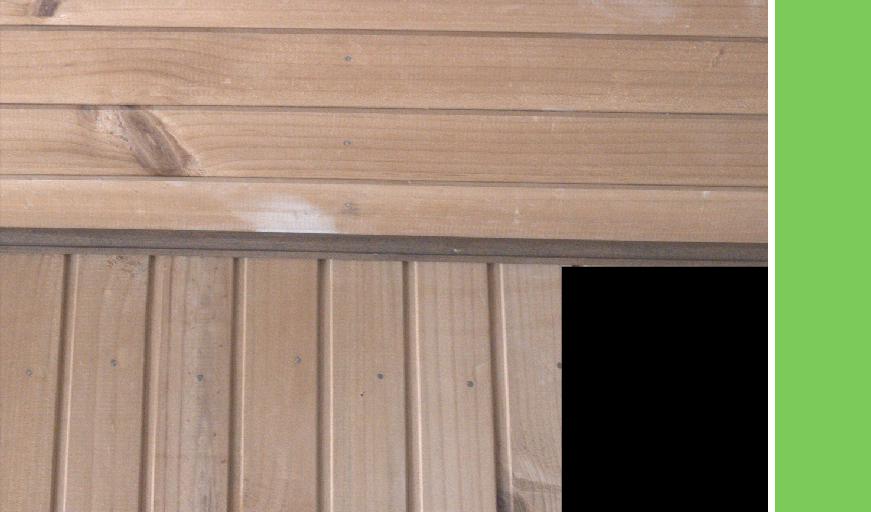}
    \vspace{-1.5em}
    \caption*{Ground Truth}
\end{subfigure}

\begin{subfigure}{\xxwidth\textwidth}
    \centering
    \includegraphics[width=0.88\columnwidth]{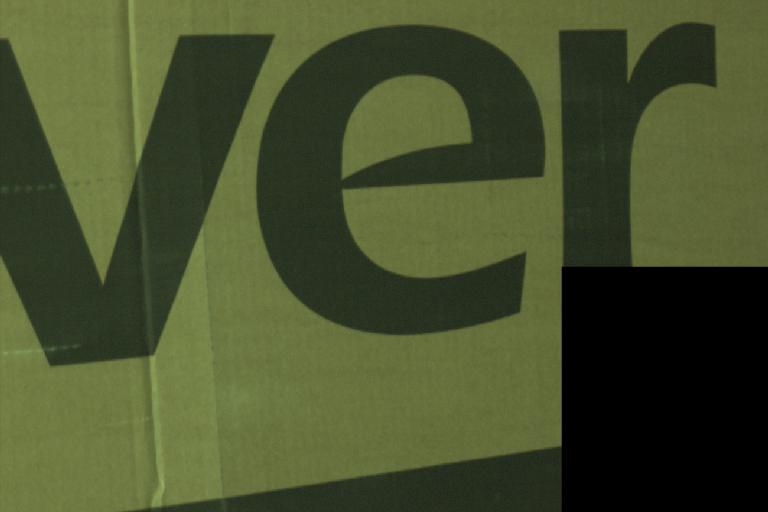}
    \vspace{-0.4em}
    \caption*{Input}
\end{subfigure}
\begin{subfigure}{\xxwidth\textwidth}
    \centering
    \includegraphics[width=\columnwidth]{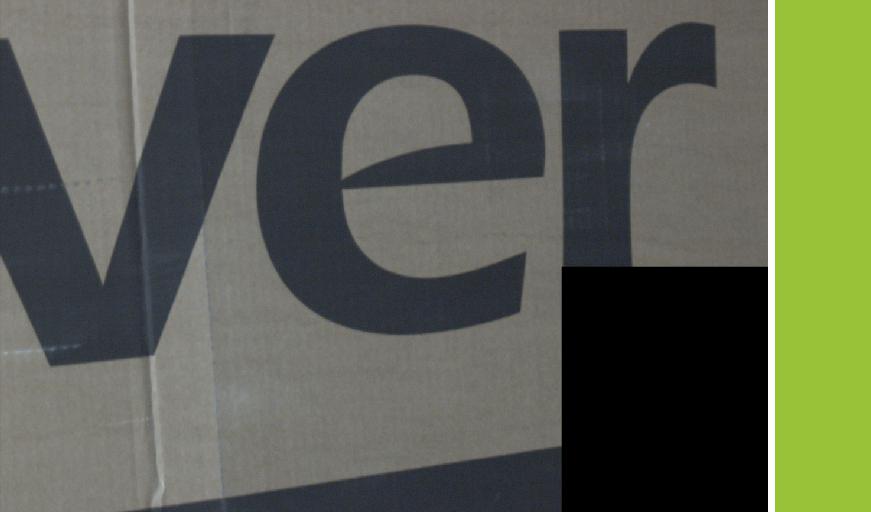}
    \vspace{-1.5em}
    \caption*{Stage1~($6.94^{\circ}$)}
\end{subfigure}
\begin{subfigure}{\xxwidth\textwidth}
    \centering
    \includegraphics[width=\columnwidth]{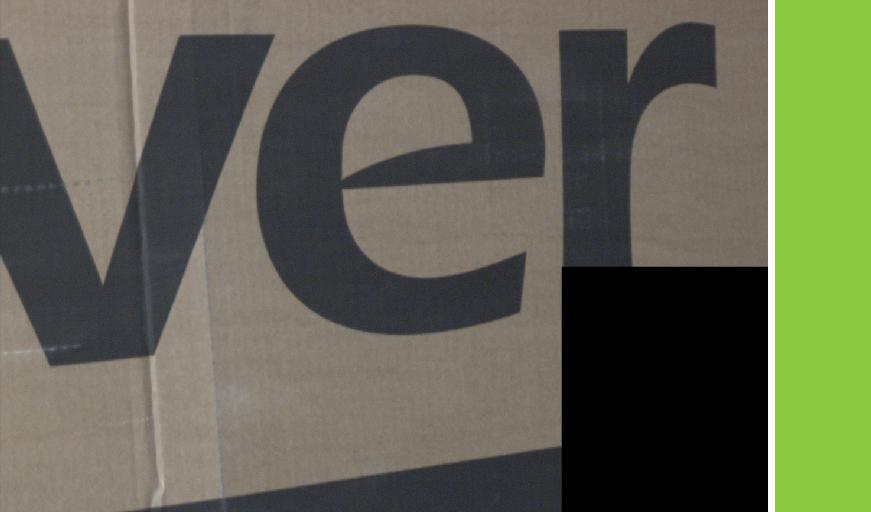}
    \vspace{-1.5em}
    \caption*{Stage2~($3.07^{\circ}$)}
\end{subfigure}
\begin{subfigure}{\xxwidth\textwidth}
    \centering
    \includegraphics[width=\columnwidth]{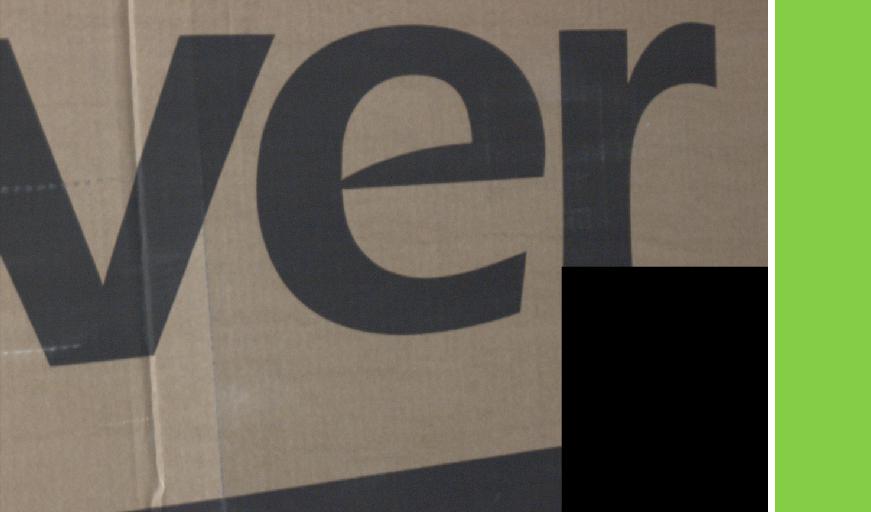}
    \vspace{-1.5em}
    \caption*{Stage3~($\mathbf{1.00^{\circ}}$)}
\end{subfigure}
\begin{subfigure}{\xxwidth\textwidth}
    \centering
    \includegraphics[width=\columnwidth]{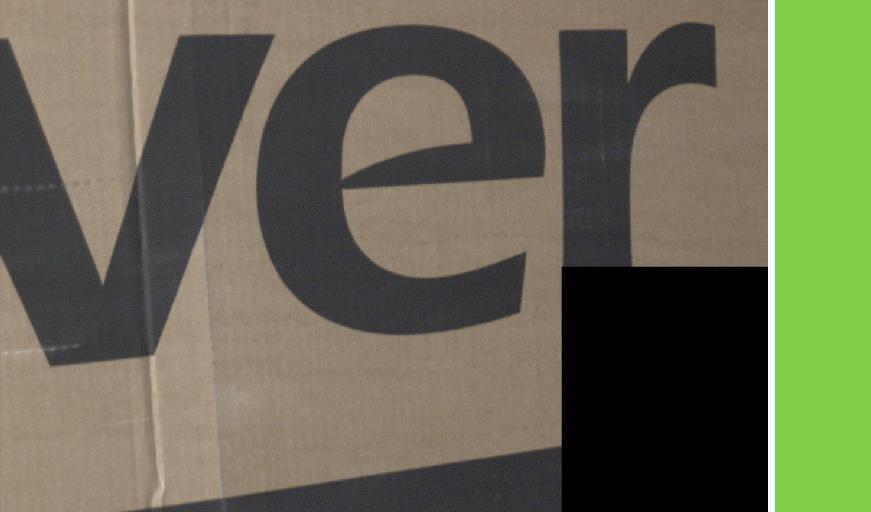}
    \vspace{-1.5em}
    \caption*{Ground Truth}
\end{subfigure}

\begin{subfigure}{\xxwidth\textwidth}
    \centering
    \includegraphics[width=0.88\columnwidth]{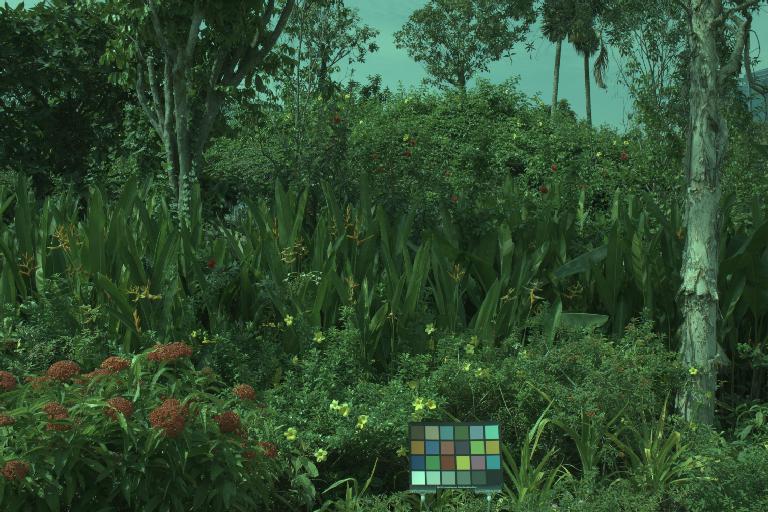}
    \vspace{-0.4em}
    \caption*{Input}
\end{subfigure}
\begin{subfigure}{\xxwidth\textwidth}
    \centering
    \includegraphics[width=\columnwidth]{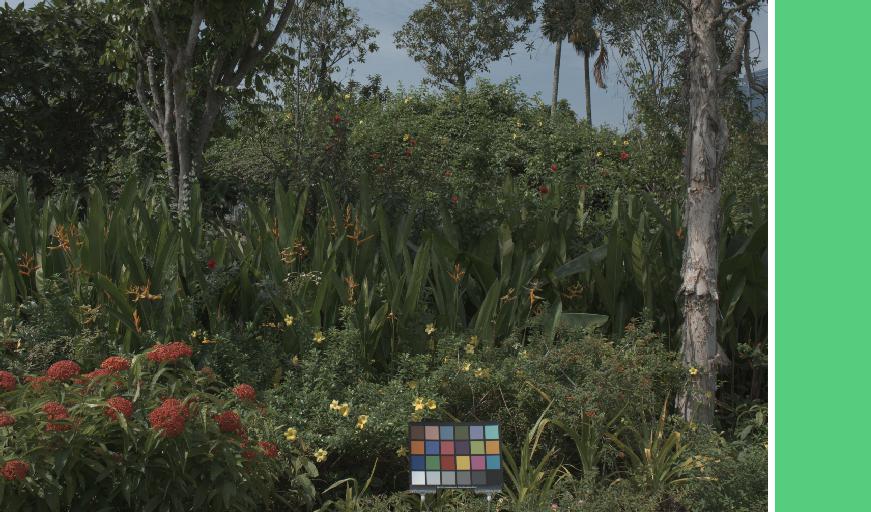}
    \vspace{-1.5em}
    \caption*{Stage1~($2.15^{\circ}$)}
\end{subfigure}
\begin{subfigure}{\xxwidth\textwidth}
    \centering
    \includegraphics[width=\columnwidth]{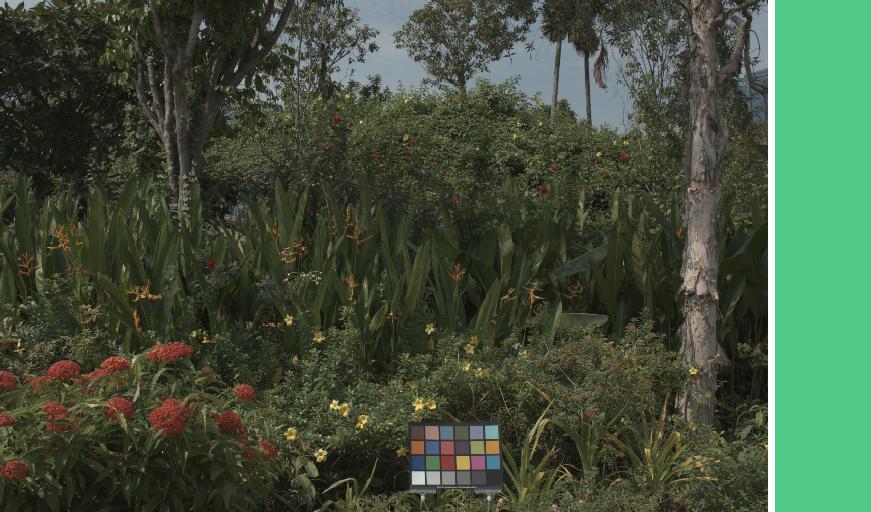}
    \vspace{-1.5em}
    \caption*{Stage2~($1.17^{\circ}$)}
\end{subfigure}
\begin{subfigure}{\xxwidth\textwidth}
    \centering
    \includegraphics[width=\columnwidth]{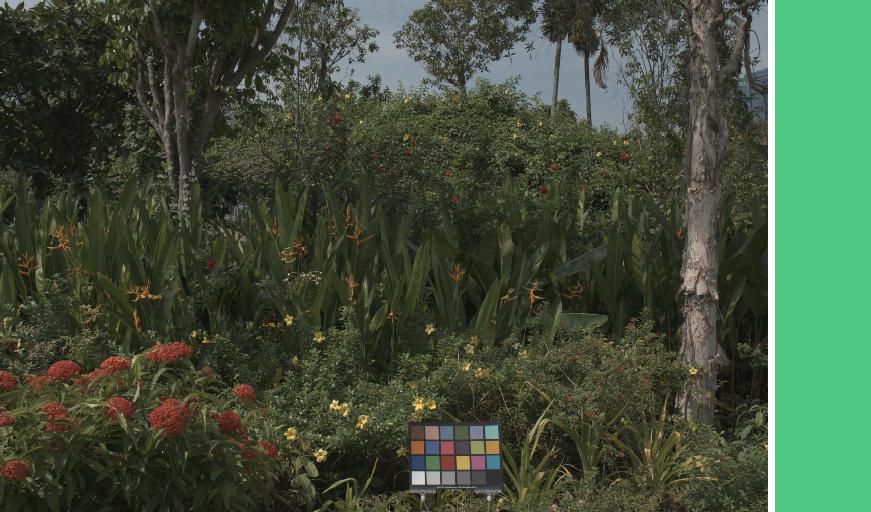}
    \vspace{-1.5em}
    \caption*{Stage3~($\mathbf{0.62^{\circ}}$)}
\end{subfigure}
\begin{subfigure}{\xxwidth\textwidth}
    \centering
    \includegraphics[width=\columnwidth]{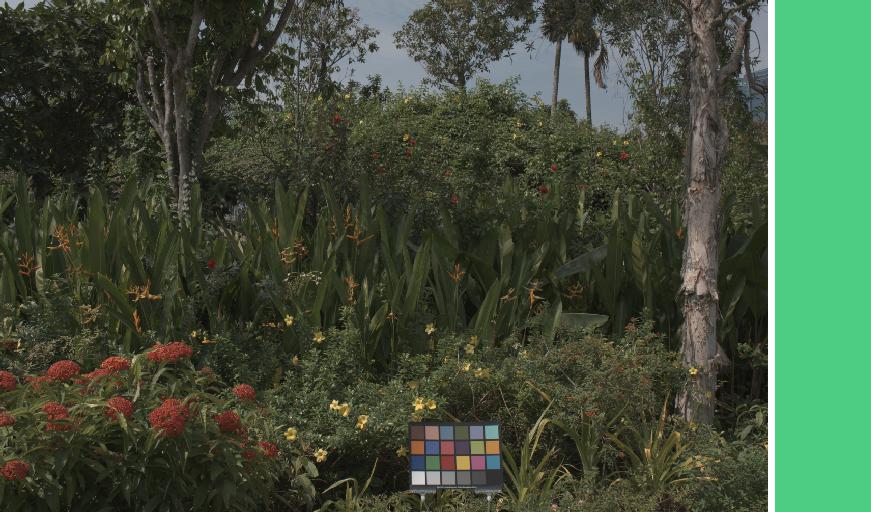}
    \vspace{-1.5em}
    \caption*{Ground Truth}
\end{subfigure}

\begin{subfigure}{\xxwidth\textwidth}
    \centering
    \includegraphics[width=0.88\columnwidth]{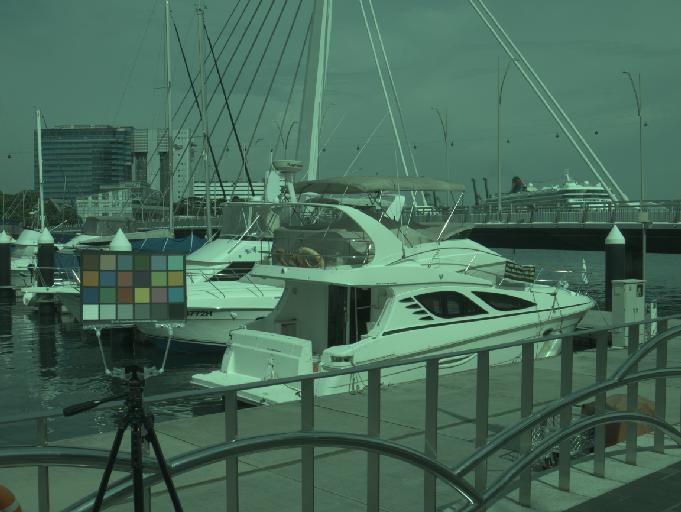}
    \vspace{-0.4em}
    \caption*{Input}
\end{subfigure}
\begin{subfigure}{\xxwidth\textwidth}
    \centering
    \includegraphics[width=\columnwidth]{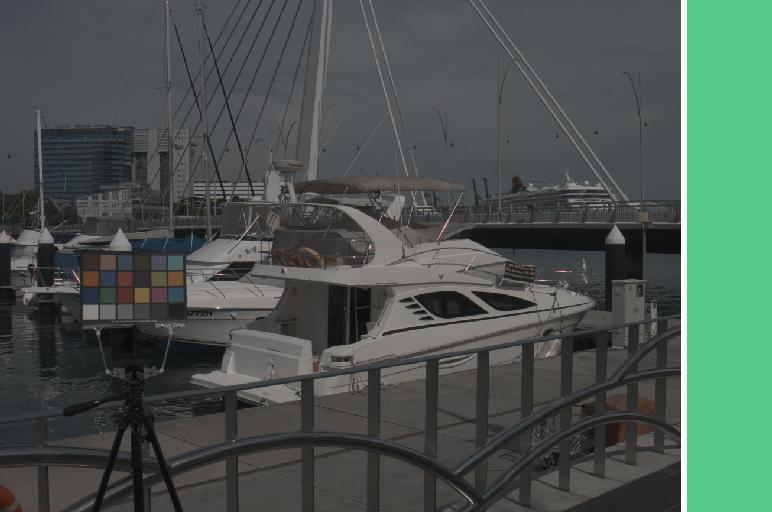}
    \vspace{-1.5em}
    \caption*{Stage1~($1.77^{\circ}$)}
\end{subfigure}
\begin{subfigure}{\xxwidth\textwidth}
    \centering
    \includegraphics[width=\columnwidth]{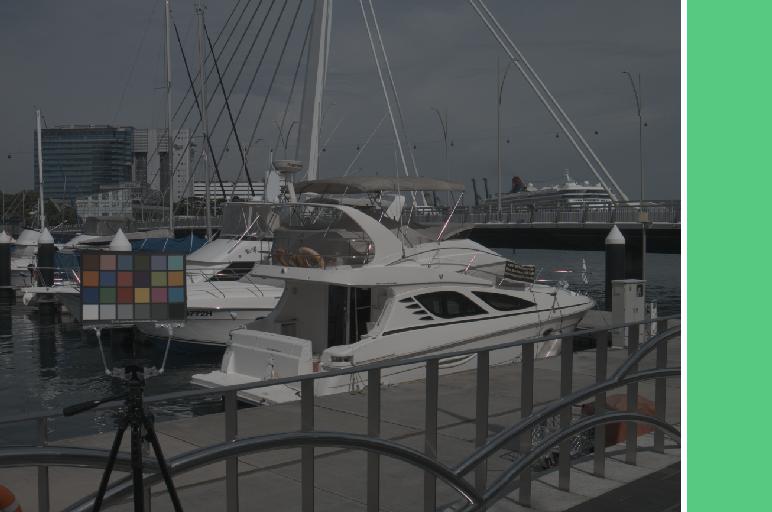}
    \vspace{-1.5em}
    \caption*{Stage2~($0.61^{\circ}$)}
\end{subfigure}
\begin{subfigure}{\xxwidth\textwidth}
    \centering
    \includegraphics[width=\columnwidth]{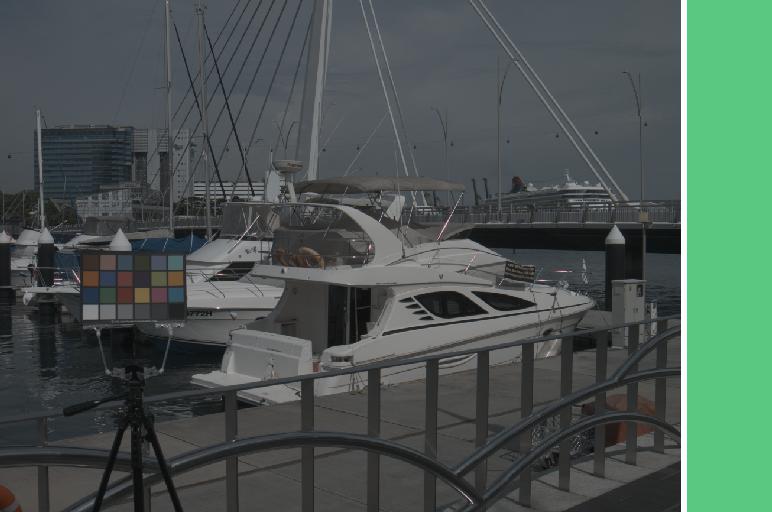}
    \vspace{-1.5em}
    \caption*{Stage3~($\mathbf{0.12^{\circ}}$)}
\end{subfigure}
\begin{subfigure}{\xxwidth\textwidth}
    \centering
    \includegraphics[width=\columnwidth]{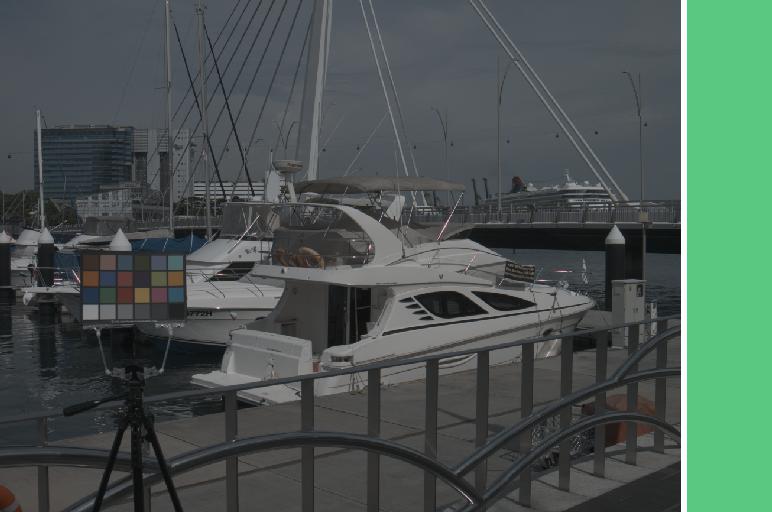}
    \vspace{-1.5em}
    \caption*{Ground Truth}
\end{subfigure}

\caption{The intermediate results in different stages of the proposed network. The images are taken from Cube-Challenge and NUS8. Note that the calibration objects are masked out when training and testing.}
\vspace{1em}
\label{fig:iteration}
\end{figure*}

\begin{table*}[!t]
    \small
    \centering
    \caption{Ablation Study of the network structure on Cube-Challenge. The best results are bold and the second best are underlined. }
   
    \begin{tabular}{ccccc|c|c|ccccc}
    \toprule
    \multicolumn{5}{c|}{Contributions} & \# Param. & FLOPs  & Mean & Med. & Tri. & Best  & Worst  \\ \cline{0-4} 
    Model & Backbone & Multi-Stages & ISAMs & Aug. &  &  & & & & 25\% & 25\% \\\hline
    $\mathit{A}$ & FC4 & unshared & - & - & 5.11M & 4.77G & 1.84 & 1.11 & 1.26 & 0.44 & 4.53  \\ 
    $\mathit{B}$ & FC4 & shared & - & - & 1.70M & 4.77G & 1.76 & 1.08 & 1.19 & 0.38 & 4.31 \\
    $\mathit{C}$ & FC4 & shared & shared & - & 1.75M & 4.77G & 1.72 & 1.09 & 1.14 & 0.33 & 4.31 \\
    $\mathit{D}$ & FC4 & shared & unshared & - & 1.84M & 4.77G & 1.66 & 0.95 & \textbf{1.03} & \textbf{0.29} & 4.29 \\
    $\mathit{E}$ & FC4 & shared & only last & - & 1.80M & 4.77G & 1.69 & 1.04 & 1.10 & 0.34 & 4.24 \\
    $\mathit{F}$ & L-FC4 & shared & unshared & - & \underline{0.82M} & \underline{3.93G} & 1.64 & \underline{0.92} & \underline{1.04} & 0.33 & 4.25 \\
    \hline
    $\mathit{G}$~(final) & L-FC4 & shared & unshared & SIE & \underline{0.82M} & \underline{3.93G} & \textbf{1.57} & \textbf{0.91} & 1.06 & \underline{0.31} & \textbf{3.92} \\ \hline
    
    $\mathit{H}$ & L-FC4-G & shared & unshared & SIE & \textbf{0.66M} & \textbf{3.87G} & 1.63 & 1.04 & 1.12 & 0.32 & \underline{3.97} \\
    $\mathit{I} $ & L-FC4-L & shared & unshared & SIE & \textbf{0.66M} & \underline{3.93G} & \underline{1.62} & 0.96 & 1.06 & 0.32 & 4.10 \\
    
    \bottomrule
    \end{tabular}
    \label{tab:ablation}
\end{table*}

\begin{figure}[t]
\centering
\includegraphics[width=1\linewidth]{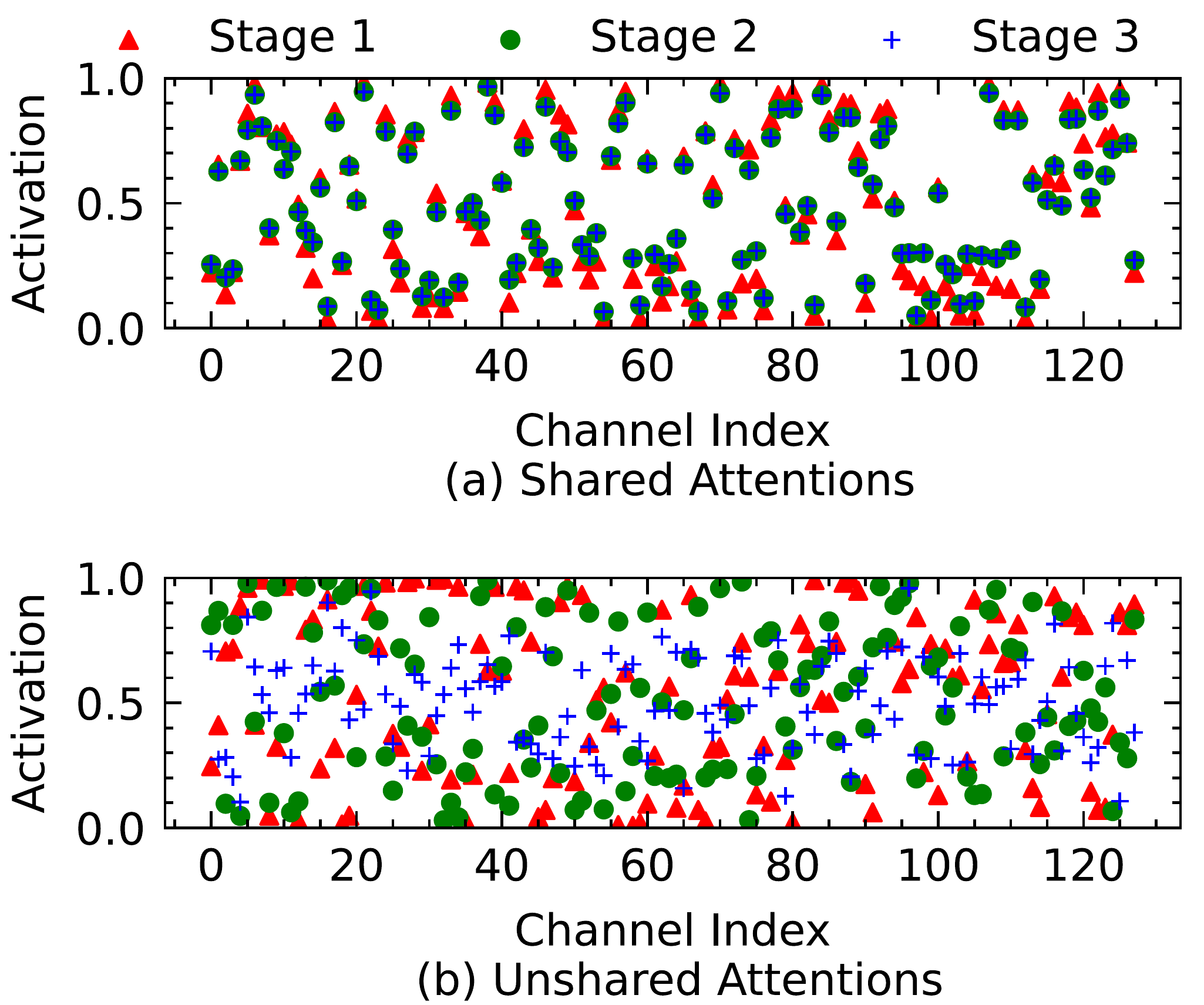}
\caption{Effect of shared/unshared attentions. (a) and~(b) show the activation distributions of the first channel re-calibration in the shared and unshared ISAMs in Model~$\mathit{C}$ and Model~$\mathit{D}$ from a single image, respectively. The unshared attentions activate more channels to learn the illuminant differences in each stage, while most activations of different stages are similar in the shared attentions. In (a), ALL of the activations from stages 2 and 3 are overlapped.}
\label{fig:att}
\end{figure}

\subsection{Ablation Study of the Network Structure} 
\label{sec:exp_net}
We have proposed a compact cascaded framework which shares the backbone in three cascaded stages and learns each stage specifically by the proposed \textit{Iteration-Specific Attention Module}~(ISAM). In this section, we do the ablation study of each component in the proposed framework on Cube-Challenge~\cite{cubechallenge}, since the NUS8 and Gehler-Shi datasets are relatively small.  
As shown in Table~\ref{tab:ablation}, we start with the baseline~(Model~$\mathit{A}$, C4~\cite{yu2020c4}, \ie,~using three unshared FC4~\cite{hu2017fc4} backbones) without the sensor-aware illuminant enhancement~(SIE). Because of the heavy network structure, directly training the cascaded network without SIE shows an unsatisfactory result compared to the model that shares the backbone in all stages~(Model~$\mathit{B}$). If we add ISAMs to Model~$\mathit{B}$ and share them in different stages~(Model~$\mathit{C}$), the network only performs slightly better on some metrics than Model~$\mathit{B}$ due to extra parameters added. Differently, when we learn the ISAMs specifically~(Model~$\mathit{D}$), all the metrics are improved significantly, which confirms the observation that the distributions in multiple stages are different and should be learnt specifically. We compare the activations of channel re-calibration in each stage between Model~$\mathit{C}$ and Model~$\mathit{D}$ in Figure~\ref{fig:att}. 
The unshared attentions show a much denser activation map in different stages while the shared attentions are insensitive to the changes of the illuminant especially in stages 2 and 3.

Besides, we also test the model when there is only the last ISAM in each stage~(Model~$\mathit{E}$). The performance of Model~$\mathit{E}$ drops a little but is still better than Model~$\mathit{B}$ and Model~$\mathit{C}$ that have either no attention or shared attentions. Then, if we replace the backbone FC4 with our dual-branch lightweight L-FC4~(Model~$\mathit{F}$), we get better overall performance than Model~$\mathit{D}$ but use only half of the parameters. And in Model~$\mathit{G}$, the usage of SIE also increases the performance of the network a lot. 

Next, we evaluate each branch of the proposed dual-branch lightweight head in Models~$\mathit{H}$~and~$\mathit{I}$~(as shown in Figure~\ref{fig:head}). Although the local branch~(L-FC4-L) and the global branch~(L-FC4-G) based models can further reduce the parameters, their performances are worse than Model~$\mathit{G}$ that has both branches. 

Finally, in Figure~\ref{fig:iteration}, we give the intermediate illuminants, corrected images and angular errors from the multiple stages~(iterations) of the proposed network to prove the effectiveness of the cascaded framework. We discover that using more stages gains much better performance than using only one stage. However, four or more stages may not improve~\cite{yu2020c4}, so in all our experiments we choose the number of stage equal to 3.
\section{Conclusion}
In this paper, we propose a novel method to learn enriched illuminants for cross and single sensor CC. First, we present self-supervised cross-sensor training by unprocessing sRGB images to the RAW domain and correcting RAW images from multiple sensors. Second, we design a compact cascaded model by sharing the backbone in a cascaded network with a novel iteration-specific attention module learnt specifically. To further reduce the parameters, we design a lightweight head with two branches for both global and local feature learning. Without training on target-sensor data, the proposed model with our training method obtains state-of-the-art performance in the cross-sensor setting. Our model trained from scratch for single-sensor CC also shows superior performance on several popular benchmarks compared with other single-sensor CC methods. Furthermore, after fine-tuning our cross-sensor model on the target sensor, its performance improves further. 



\bibliographystyle{IEEEtran}
\bibliography{egbib.bib}
%

%




\end{document}